\documentclass[10pt,twocolumn,letterpaper]{article}

\usepackage[pagenumbers]{noname}
\usepackage{algorithm}
\usepackage{algpseudocode}
\usepackage{multirow}
\usepackage{booktabs}
\usepackage{fontawesome5}
\usepackage[table]{xcolor}
\usepackage{colortbl}
\usepackage{graphicx}
\usepackage{lipsum}
\usepackage{pifont}
\usepackage{xparse}
\usepackage{pgf}
\usepackage{float}
\usepackage{makecell}

\definecolor{heatred}{RGB}{248,105,107}
\definecolor{heatyellow}{RGB}{255,235,132}
\definecolor{heatgreen}{RGB}{99,190,123}

\NewDocumentCommand{\RankCell}{O{} m m}{%
  \pgfmathsetmacro{\score}{100 * (10 - #3) / 9.0}%
  \pgfmathtruncatemacro{\s}{round(\score)}%
  
  \ifnum\s<50
    \pgfmathtruncatemacro{\mix}{100-2*\s}%
    \edef\heatcol{heatred!\mix!heatyellow}%
  \else
    \pgfmathtruncatemacro{\mix}{200-2*\s}%
    \edef\heatcol{heatyellow!\mix!heatgreen}%
  \fi
  
  \edef\heatcolsoft{\heatcol!45!white}%
  \expandafter\cellcolor\expandafter{\heatcolsoft}%
  
  \if\relax\detokenize{#1}\relax #2\else #1\fi
}

\NewDocumentCommand{\RankCellTwelve}{O{} m m}{%
  \pgfmathsetmacro{\score}{100 * (12 - #3) / 11.0}%
  \pgfmathtruncatemacro{\s}{round(\score)}%
  
  \ifnum\s<50
    \pgfmathtruncatemacro{\mix}{100-2*\s}%
    \edef\heatcol{heatred!\mix!heatyellow}%
  \else
    \pgfmathtruncatemacro{\mix}{200-2*\s}%
    \edef\heatcol{heatyellow!\mix!heatgreen}%
  \fi
  
  \edef\heatcolsoft{\heatcol!45!white}%
  \expandafter\cellcolor\expandafter{\heatcolsoft}%
  
  \if\relax\detokenize{#1}\relax #2\else #1\fi
}

\NewDocumentCommand{\RankCellEleven}{O{} m m}{%
  \pgfmathsetmacro{\score}{100 * (11 - #3) / 10.0}%
  \pgfmathtruncatemacro{\s}{round(\score)}%
  \ifnum\s<50
    \pgfmathtruncatemacro{\mix}{100-2*\s}%
    \edef\heatcol{heatred!\mix!heatyellow}%
  \else
    \pgfmathtruncatemacro{\mix}{200-2*\s}%
    \edef\heatcol{heatyellow!\mix!heatgreen}%
  \fi
  \edef\heatcolsoft{\heatcol!45!white}%
  \expandafter\cellcolor\expandafter{\heatcolsoft}%
  \if\relax\detokenize{#1}\relax #2\else #1\fi
}

\NewDocumentCommand{\RankCellTen}{O{} m m}{%
  \pgfmathsetmacro{\score}{100 * (10 - #3) / 9.0}%
  \pgfmathtruncatemacro{\s}{round(\score)}%
  \ifnum\s<50
    \pgfmathtruncatemacro{\mix}{100-2*\s}%
    \edef\heatcol{heatred!\mix!heatyellow}%
  \else
    \pgfmathtruncatemacro{\mix}{200-2*\s}%
    \edef\heatcol{heatyellow!\mix!heatgreen}%
  \fi
  \edef\heatcolsoft{\heatcol!45!white}%
  \expandafter\cellcolor\expandafter{\heatcolsoft}%
  \if\relax\detokenize{#1}\relax #2\else #1\fi
}

\NewDocumentCommand{\RankCellSix}{O{} m m}{%
  \pgfmathsetmacro{\score}{100 * (6 - #3) / 5.0}%
  \pgfmathtruncatemacro{\s}{round(\score)}%
  
  \ifnum\s<50
    \pgfmathtruncatemacro{\mix}{100-2*\s}%
    \edef\heatcol{heatred!\mix!heatyellow}%
  \else
    \pgfmathtruncatemacro{\mix}{200-2*\s}%
    \edef\heatcol{heatyellow!\mix!heatgreen}%
  \fi
  
  \edef\heatcolsoft{\heatcol!45!white}%
  \expandafter\cellcolor\expandafter{\heatcolsoft}%
  
  \if\relax\detokenize{#1}\relax #2\else #1\fi
}

\NewDocumentCommand{\RankCellThree}{O{} m m}{%
  \pgfmathsetmacro{\score}{100 * (3 - #3) / 2.0}%
  \pgfmathtruncatemacro{\s}{round(\score)}%
  
  \ifnum\s<50
    \pgfmathtruncatemacro{\mix}{100-2*\s}%
    \edef\heatcol{heatred!\mix!heatyellow}%
  \else
    \pgfmathtruncatemacro{\mix}{200-2*\s}%
    \edef\heatcol{heatyellow!\mix!heatgreen}%
  \fi
  
  \edef\heatcolsoft{\heatcol!45!white}%
  \expandafter\cellcolor\expandafter{\heatcolsoft}%
  
  \if\relax\detokenize{#1}\relax #2\else #1\fi
}

\NewDocumentCommand{\RankCellFour}{O{} m m}{%
  \pgfmathsetmacro{\score}{100 * (4 - #3) / 3.0}%
  \pgfmathtruncatemacro{\s}{round(\score)}%
  
  \ifnum\s<50
    \pgfmathtruncatemacro{\mix}{100-2*\s}%
    \edef\heatcol{heatred!\mix!heatyellow}%
  \else
    \pgfmathtruncatemacro{\mix}{200-2*\s}%
    \edef\heatcol{heatyellow!\mix!heatgreen}%
  \fi
  
  \edef\heatcolsoft{\heatcol!45!white}%
  \expandafter\cellcolor\expandafter{\heatcolsoft}%
  
  \if\relax\detokenize{#1}\relax #2\else #1\fi
}

\NewDocumentCommand{\RankCellTwo}{O{} m m}{%
  \pgfmathsetmacro{\score}{100 * (2 - #3) / 1.0}%
  \pgfmathtruncatemacro{\s}{round(\score)}%
  
  \ifnum\s<50
    \pgfmathtruncatemacro{\mix}{100-2*\s}%
    \edef\heatcol{heatred!\mix!heatyellow}%
  \else
    \pgfmathtruncatemacro{\mix}{200-2*\s}%
    \edef\heatcol{heatyellow!\mix!heatgreen}%
  \fi
  
  \edef\heatcolsoft{\heatcol!45!white}%
  \expandafter\cellcolor\expandafter{\heatcolsoft}%
  
  \if\relax\detokenize{#1}\relax #2\else #1\fi
}

\NewDocumentCommand{\RankCellFive}{O{} m m}{%
  \pgfmathsetmacro{\score}{100 * (5 - #3) / 4.0}%
  \pgfmathtruncatemacro{\s}{round(\score)}%
  
  \ifnum\s<50
    \pgfmathtruncatemacro{\mix}{100-2*\s}%
    \edef\heatcol{heatred!\mix!heatyellow}%
  \else
    \pgfmathtruncatemacro{\mix}{200-2*\s}%
    \edef\heatcol{heatyellow!\mix!heatgreen}%
  \fi
  
  \edef\heatcolsoft{\heatcol!45!white}%
  \expandafter\cellcolor\expandafter{\heatcolsoft}%
  
  \if\relax\detokenize{#1}\relax #2\else #1\fi
}

\NewDocumentCommand{\RankCellEight}{O{} m m}{%
  \pgfmathsetmacro{\score}{100 * (8 - #3) / 7.0}%
  \pgfmathtruncatemacro{\s}{round(\score)}%

  \ifnum\s<50
    \pgfmathtruncatemacro{\mix}{100-2*\s}%
    \edef\heatcol{heatred!\mix!heatyellow}%
  \else
    \pgfmathtruncatemacro{\mix}{200-2*\s}%
    \edef\heatcol{heatyellow!\mix!heatgreen}%
  \fi

  \edef\heatcolsoft{\heatcol!45!white}%
  \expandafter\cellcolor\expandafter{\heatcolsoft}%

  \if\relax\detokenize{#1}\relax #2\else #1\fi
}

\definecolor{nonameblue}{rgb}{0.21,0.49,0.74}
\usepackage[pagebackref,breaklinks,colorlinks,allcolors=nonameblue]{hyperref}

\title{Adaptive Densification for High-Fidelity and Efficient Sparse Gaussian Splatting in Arbitrary-Scale Super-Resolution}

\author{
Giulio Federico$^{1,2}$\thanks{Corresponding author.} \quad
Giuseppe Amato$^{2}$ \quad
Claudio Gennaro$^{2}$ \quad
Fabio Carrara$^{2}$ \quad
Marco Di Benedetto$^{2}$ \\
\\
$^1$University of Pisa, Pisa, Italy \quad
$^2$ISTI-CNR, Pisa, Italy \\
{\tt\small \{giulio.federico, giuseppe.amato, claudio.gennaro, fabio.carrara, marco.dibenedetto\}@isti.cnr.it}
}

\begin{document}
\maketitle
\begin{abstract}

Arbitrary-Scale Super-Resolution (ASR) aims to reconstruct high-resolution images at any continuous magnification. While 2D Gaussian Splatting (GS) has recently shown great promise for ASR, current methods struggle to balance visual quality and computational cost. Approaches targeting high fidelity rely on powerful backbones and uniform, highly dense Gaussian grids, leading to prohibitive memory and inference costs. Conversely, methods prioritizing efficiency aggressively simplify their architectures, severely compromising visual quality. To bridge this gap, we observe that a core capability of GS remains largely underexplored in ASR: the potential for dynamic densification, i.e., the spatially adaptive allocation of Gaussians based on image content. Unlike standard scene fitting, where densification is guided by a known ground truth, applying this to ASR is highly non-trivial because the high-resolution target is exactly what the model must predict. To address this challenge, we propose QuADA-GS, an approach that retains a powerful representational backbone but autonomously predicts where to allocate Gaussians relying strictly on the low-resolution input. By adopting a sparse approach, QuADA-GS refines features and increases Gaussian density strictly where structural complexity demands it. Because this adaptive allocation produces a non-uniform hierarchical topology, we introduce a novel, highly efficient communication mechanism to process these sparse features, bypassing standard dense bottlenecks. Extensive experiments indicate that our approach successfully balances visual quality and computational requirements, providing an improved and competitive trade-off for ASR.
\end{abstract}

\section{Introduction}
\label{sec:intro}

Super-resolution (SR) methods have achieved remarkable reconstruction quality, yet most approaches operate at fixed integer scale factors, an assumption that rarely holds in production pipelines. 
To address this, Arbitrary-Scale SR emerged. While Implicit Neural Representations (INRs)~\cite{chen2021learning, hu2019meta, wang2021learning,lee2022local, Yao_2023_CVPR, cao2023ciaosr, Wei_2023_CVPR, Chen_2023_CVPR, NEURIPS2021_6e7d5d25,He_2024_CVPR} advanced ASR by decoding continuous coordinates via Multi-Layer Perceptrons (MLPs), exhaustive pixel-by-pixel querying causes severe computational bottlenecks at high resolutions, often resulting in oversmoothed outputs that fail to capture sharp, high-frequency details.

2D Gaussian Splatting \cite{kerbl20233d} has emerged as a powerful continuous representation. Recent ASR adaptations \cite{hu2025gaussiansr, Chen_2025_ICCV, grape, continuoussr} leverage GS to enhance visual quality and bypass exhaustive pixel inference. However, \textit{current approaches face a severe trade-off between reconstruction quality and computational costs} dictated by a rigid, uniform processing philosophy. High-fidelity methods rely on powerful backbones and heavy upsampling modules to uniformly allocate a massive number of Gaussians, incurring prohibitive overhead. Conversely, efficient alternatives discard powerful backbones to reduce costs. While this maintains low memory, it severely degrades quality because the weaker network cannot unveil missing high-resolution details from the LR input. Ultimately, by distributing a \textit{uniform density of Gaussians} across both flat areas and complex details, these strategies leave a core strength of GS largely underexplored: \textit{dynamic densification}. In standard GS, primitives are dynamically allocated using gradients derived from the data to be fitted. Adapting this mechanism to ASR is highly non-trivial: without the high-resolution target during inference, traditional gradient-guided densification cannot be applied.


To this end, we introduce \textbf{Qu}adtree \textbf{Ada}ptive \textbf{G}aussian \textbf{S}platting (QuADA-GS). Our framework retains powerful feature extractors on the Low-Resolution (LR) input, preserving the crucial representation capacity needed to uncover latent high-frequency details. However, we eliminate the severe memory bottlenecks caused by uniformly upsampling the entire feature map to high resolutions. Instead of a dense and uniform allocation, QuADA-GS dynamically evolves the base feature map into a sparse quadtree structure, keeping computational and memory costs strictly bounded by refining only the regions that truly require it. Since this sparse architecture places features at different spatial resolutions and tree depths, we introduce the \textit{Hierarchical Pointer Convolution} to let them communicate seamlessly. This operator enables each feature to locate its topological neighbors via an $\mathcal{O}(1)$ search, ensuring efficient feature aggregation while entirely bypassing dense operational bottlenecks.
\section{Related Work}


\paragraph{Data and Evaluation.}
SR aims to recover a High-Resolution (HR) image from the LR input. Early benchmarks used clean, synthetic LR-HR pairs, but the field shifted toward composite, real-world degradations for Blind SR~\cite{wang2021real, Zhang_2021_ICCV} and perceptually-guided ground truths~\cite{chen2023human, chen2025toward}, since HR references often contain flaws that modern generative models surpass in perceived quality.

\begin{figure*}[t]
\centering
\includegraphics[width=0.94\textwidth]{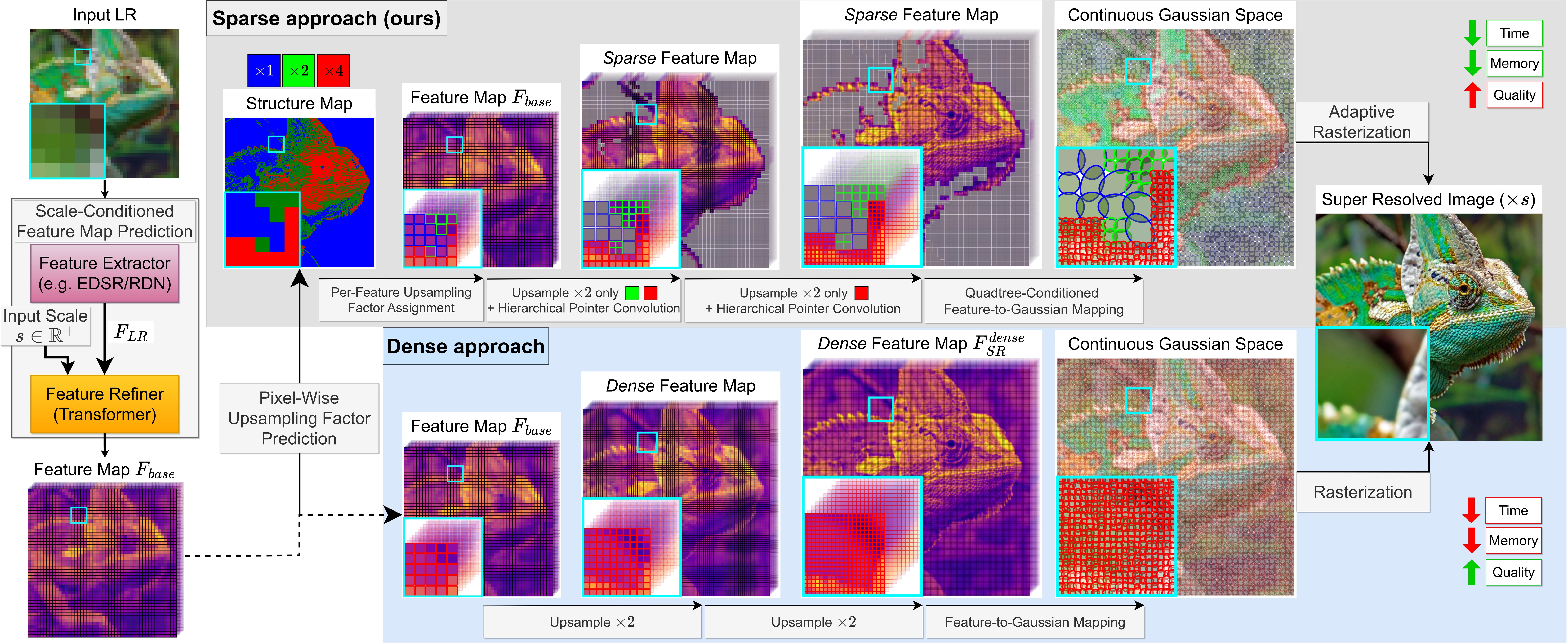}
\caption{\textbf{Dense vs.\ proposed sparse GS-ASR:} Both extract scale-conditioned $\mathbf{F}_{base}$ from $I_{LR}$. The dense pipeline (cyan) uniformly upsamples and rasterizes for SR, allocating equal Gaussian density regardless of local complexity. In contrast, our Structure Map predicts independent per-feature expansion, creating a sparse quadtree topology. Features communicate across this structure via Hierarchical Pointer Convolution. They are then decoded into Gaussians, sized by quadtree level (large for flat areas, small for details), and rasterized for SR.}
\label{fig:dense_vs_sparse}
\end{figure*}


\paragraph{Traditional SR and Generative Hallucinations.}
Traditional architectures advanced through deep convolutional networks~\cite{Kim_2016_CVPR, kim2016deeply, lai2017deep}, efficient residual designs~\cite{dong2016accelerating, wang2018esrgan, zhang2018image, zhang2018residual, Li_2019_CVPR}, and window-based Transformers~\cite{Liang_2021_ICCV, zhang2022efficient, chen2023activating, zhang2024transcending}, yet remain constrained to fixed integer scales. Recently, diffusion models extended synthesis to ASR by embedding Implicit Neural Representations into the denoising process~\cite{Gao_2023_CVPR} or a latent-diffusion decoder~\cite{Kim_2024_CVPR}. However, heavily relying on massive web-scale priors~\cite{yu2024scaling}, these iterative models struggle to guarantee faithful reconstruction. Despite efforts aligning generative and pixel-space signals~\cite{yi2025fine}, they act as generative samplers, not strict reconstructors, frequently hallucinating structural details absent from the LR input~\cite{chen2024ssl, wu2024seesr, ren2025hallucination}.

\paragraph{Arbitrary-Scale SR and Gaussian Splatting.}
To maintain structural fidelity without being constrained by fixed grids, ASR frames SR as a continuous representation problem. Meta-SR~\cite{hu2019meta} and scale-aware plug-ins~\cite{wang2021learning} broke the integer-scale constraint by dynamically predicting filter weights. LIIF~\cite{chen2021learning} formalized ASR by training an MLP to predict RGB values from continuous spatial coordinates, a paradigm also effective for preserving sharp text in screen content (ITSRN~\cite{NEURIPS2021_6e7d5d25}). To overcome the MLP's tendency to suppress high frequencies, LTE~\cite{lee2022local} mapped textures into 2D Fourier space, while LINF~\cite{Yao_2023_CVPR} modeled local texture distributions via Normalizing Flows. To expand the receptive field, CiaoSR~\cite{cao2023ciaosr} introduced implicit attention-in-attention. SRNO~\cite{Wei_2023_CVPR} reframed ASR through neural operators to map continuous function spaces, while LIT~\cite{Chen_2023_CVPR} improved inputs with frequency encodings. Since querying an MLP for every target coordinate is highly memory and time intensive, LMF~\cite{He_2024_CVPR} introduced a decoupled paradigm, confining heavy computation to the LR latent space and using a lightweight dynamic renderer.


Recently, 2D Gaussian Splatting~\cite{kerbl20233d} has been adapted for ASR, but existing methods enforce a rigid trade-off between representational power and cost while treating every spatial location identically. GaussianSR~\cite{hu2025gaussiansr} pairs a lightweight backbone with one Gaussian kernel per low-resolution pixel, chosen from a fixed bank, computing continuous high-resolution values by alpha-blending neighboring LR features and restricting this to a channel subset, bicubically interpolating the rest; lacking a CUDA rasterizer, it stays slow and memory-heavy at every scale. GRAPE~\cite{grape}, in contrast, predicts one anisotropic Gaussian per LR pixel through a single point-wise layer, trading fidelity for genuine real-time rendering. ContinuousSR~\cite{continuoussr} pairs a powerful backbone with a Deep Gaussian Prior that statistically regularizes a fixed four Gaussians per pixel. GSASR~\cite{Chen_2025_ICCV} pushes fidelity further, updating learnable Gaussian embeddings via cross- and self-attention and densifying them to 16 Gaussians per pixel through cascaded pixel shuffles and convolutions, with MLPs predicting continuous Gaussian parameters for custom CUDA rendering, at the cost of heavy memory and runtime overhead.

Existing GS-based ASR methods trade efficiency (lightweight backbones, poor reconstruction quality) for high-fidelity (powerful backbones, prohibitive memory/time costs). Crucially, \textit{all densify uniformly} across flat and textured regions, with a \textit{fixed number} assigned per pixel, ignoring GS's core strength: \textit{content-adaptive densification}.

\paragraph{Content-adaptive SR.}
Content-adaptive methods outside GS (e.g., ClassSR~\cite{classSR}, PCSR~\cite{pcsr}, QDM~\cite{qdm}) operate \textit{early} by routing entire LR patches to distinct branches based on estimated complexity. This coarse approach is suboptimal: complexity varies per-pixel, isolating patches hinders global communication (often causing boundary artifacts), and the final output remains a fixed-size grid. Instead, we apply a strong global feature extractor first and shift adaptive refinement \textit{downstream} to act strictly per-feature.

Furthermore, existing sparse representations struggle with the resulting non-uniform topology. Hash grids~\cite{instantNGP} require per-scene optimization, while standard sparse convolutions~\cite{SubmanifoldSparseConvNet,Choy_2019_CVPR} and tree-based networks~\cite{o_cnn, Jayaraman_2018_ECCV,Chitta_2020_WACV, octformer} typically restrict operations to fixed depths, failing to efficiently mix adjacent features across varying refinement levels. In contrast, since our adaptive densification builds a deterministic quadtree, we exploit this structural regularity \textit{a priori} to design a specialized and highly efficient cross-depth communication mechanism for ASR.
\section{Method}
\label{sec:method}

\subsection{Preliminaries: Dense vs.\ Sparse GS-based ASR}
\label{sec:preliminaries}

Given $I_{LR}\in\mathbb{R}^{3\times H\times W}$ and a target scale $s\in\mathbb{R}^+$, reconstructing the super-resolved image $\mathbf{I}_{SR}\in\mathbb{R}^{3\times H_{SR}\times W_{SR}}$, with $H_{SR}=\lfloor Hs\rfloor$ and $W_{SR}=\lfloor Ws\rfloor$, requires a feature extractor capable of revealing, from $I_{LR}$ alone, a feature map $\mathbf{F}_{base}\in\mathbb{R}^{C\times H\times W}$ (where $C$ denotes the number of feature channels) that already contains all information needed to reconstruct $I_{SR}$. While this extraction phase heavily impacts the computational budget, drastically trimming the backbone irreversibly limits recoverable details, regardless of the subsequent Gaussian decoding. To optimize the quality-efficiency trade-off, we seek computational savings elsewhere, maintaining a full-strength extractor and adopting the same \textit{initial} architecture as GSASR~\cite{Chen_2025_ICCV}, the current state of the art in reconstruction quality.

Concretely, an off-the-shelf feature extractor (EDSR-Baseline~\cite{Lim_2017_CVPR_Workshops}, RDN~\cite{zhang2018residual}) first produces an initial feature map $\mathbf{F}_{LR}\in\mathbb{R}^{D\times H\times W}$ of channel depth $D$ from $I_{LR}$. 
Then, a single $K\times K$ grid of learnable embeddings $\mathbf{E}_{gs}$ is introduced, where $K$ defines the spatial dimension of this base grid. To match the spatial resolution of the feature map, this grid is spatially replicated  to match the resolution of $\mathbf{F}_{LR}$. This expansion produces a tensor $\tilde{\mathbf{E}}_{gs}\in\mathbb{R}^{C\times H\times W}$, where every $K\times K$ spatial window is initially identical to the others. A Transformer feature (refiner) then breaks this symmetry, giving each tile a unique, scale- and content-aware identity in two stages. First, each tile is modulated with a scale embedding $\mathbf{e}_s=\mathbf{W}_s(1/s)\in\mathbb{R}^C$ (a linear projection of the inverse scale $1/s$) via Multi-Head Attention (MHA), and made to look at its own window of $\mathbf{F}_{LR}$, denoted $\mathbf{F}_{LR}^{win}$, via Window Cross-Attention (W-MCA) \cite{swin_transformer}, so tiles absorb local LR content:
\begin{equation}
    \mathbf{E}_{gs}^{\text{ctx}} = \text{W-MCA}\big(\text{MHA}(\tilde{\mathbf{E}}_{gs}, \mathbf{e}_s),\, \mathbf{F}_{LR}^{win}\big).
\end{equation}
Second, tiles are re-modulated with $\mathbf{e}_s$ via MHA and made to talk to each other via a shifted Window Multi-head Self-Attention (SW-MSA), so neighboring tiles reconcile their boundaries before decoding:
\begin{equation}
    \mathbf{F}_{base} = \text{SW-MSA}\big(\text{MHA}(\mathbf{E}_{gs}^{\text{ctx}}, \mathbf{e}_s)\big).
\end{equation}
The result, $\mathbf{F}_{base}\in\mathbb{R}^{C\times H\times W}$, is thus a feature map aware of both the target scale and its own spatial neighborhood (Fig. \ref{fig:dense_vs_sparse}), at a cost proportional to the LR resolution.

\paragraph{Where our sparse approach diverges from the dense one.}

\begin{figure}[t]
\centering
\includegraphics[width=1.0\columnwidth]{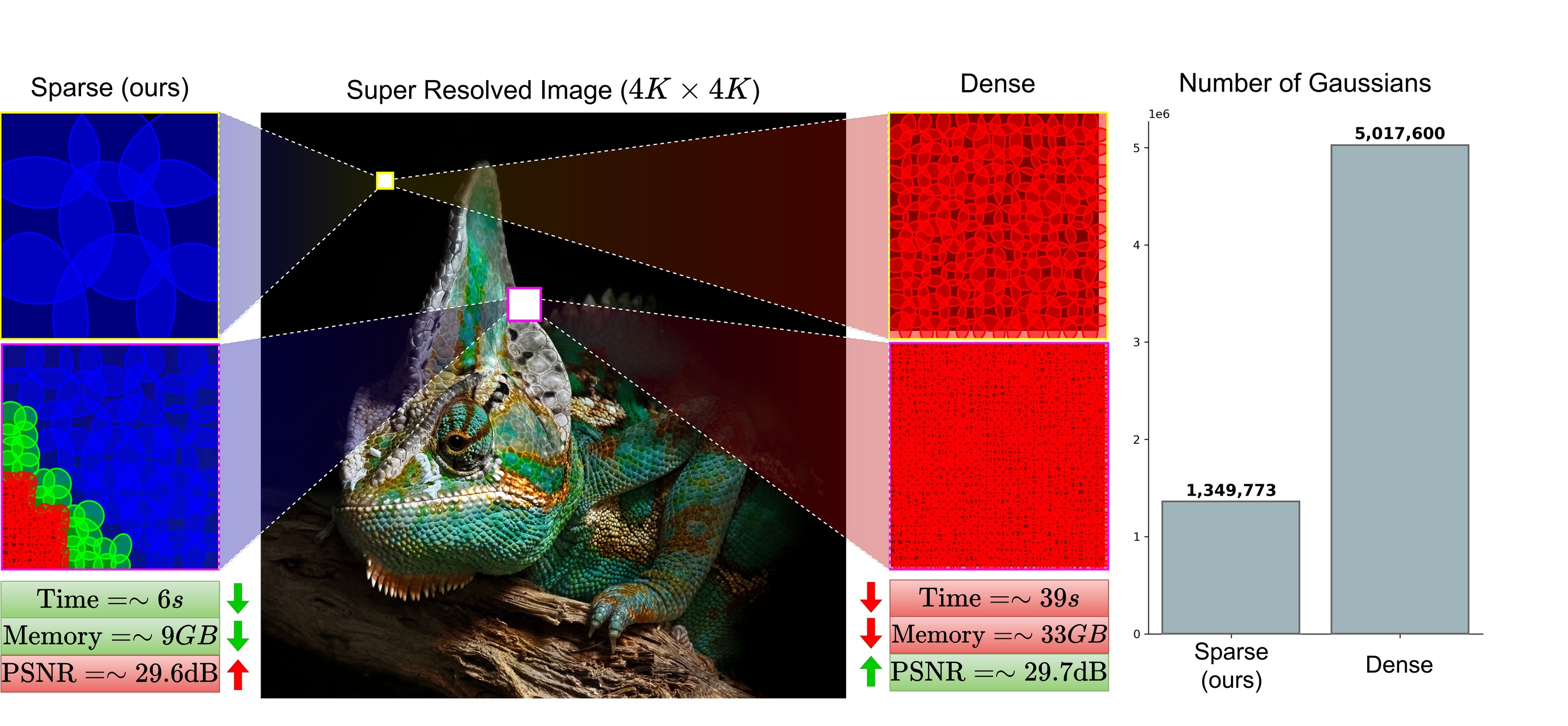}
\caption{\textbf{Quality-cost trade-off:} For $\times 8$ SR (500px to $8K$), our sparse method uses $\sim$1M Gaussians vs.\ $\sim$5M (dense), cutting time by $\sim$85\% and memory by $\sim$73\% for a mere $\sim$0.1dB quality drop.}
\label{fig:example_dense_vs_sparse}
\end{figure}

A dense approach uniformly upsamples $\mathbf{F}_{base}$ before decoding: GSASR~\cite{Chen_2025_ICCV} expands it into $\mathbf{F}_{SR}^{dense}\in\mathbb{R}^{C\times 4H\times 4W}$ via convolutions and Pixel Shuffle~\cite{Shi_2016_CVPR}, decoding all $16HW$ features into 2D Gaussians (Fig.~\ref{fig:dense_vs_sparse}, cyan) regardless of content, scaling poorly with input size. Our sparse approach (Fig.~\ref{fig:dense_vs_sparse}, gray) instead predicts a \textit{Structure Map} $\in\mathbb{Z}^{H\times W}$ (Sec.~\ref{sec:structure_map}) assigning each $\mathbf{F}_{base}$ feature an expansion factor ($\times1$, $\times2$, $\times4$), refined into a quadtree topology. A \textit{Hierarchical Pointer Convolution} (Sec.~\ref{sec:hpc}) allows multi-level feature communication before decoding into 2D Gaussians. By adapting rasterized sizes to their quadtree levels (Sec.~\ref{sec:rasterization}), large Gaussians cover areas dense approaches tile redundantly. In Fig.~\ref{fig:example_dense_vs_sparse}, this reduces Gaussians by $\sim$80\%, inference time by $\sim$85\%, and memory by $\sim$73\%, sacrificing only $\sim$0.1dB in quality.

\subsection{Determining the Expansion Factor per Feature}
\label{sec:structure_map}

Given $\mathbf{F}_{base}\in\mathbb{R}^{C\times H\times W}$, our sparse approach assigns each feature $f_i\in\mathbb{R}^C$ an expansion factor ($\times 1, \times 2, \times 4$). Following GSASR~\cite{Chen_2025_ICCV}, we set $\times4$ as the maximum factor and once decoded into 2D Gaussians they are rasterized.

\begin{figure}[H]
\centering
\includegraphics[width=1.0\columnwidth]{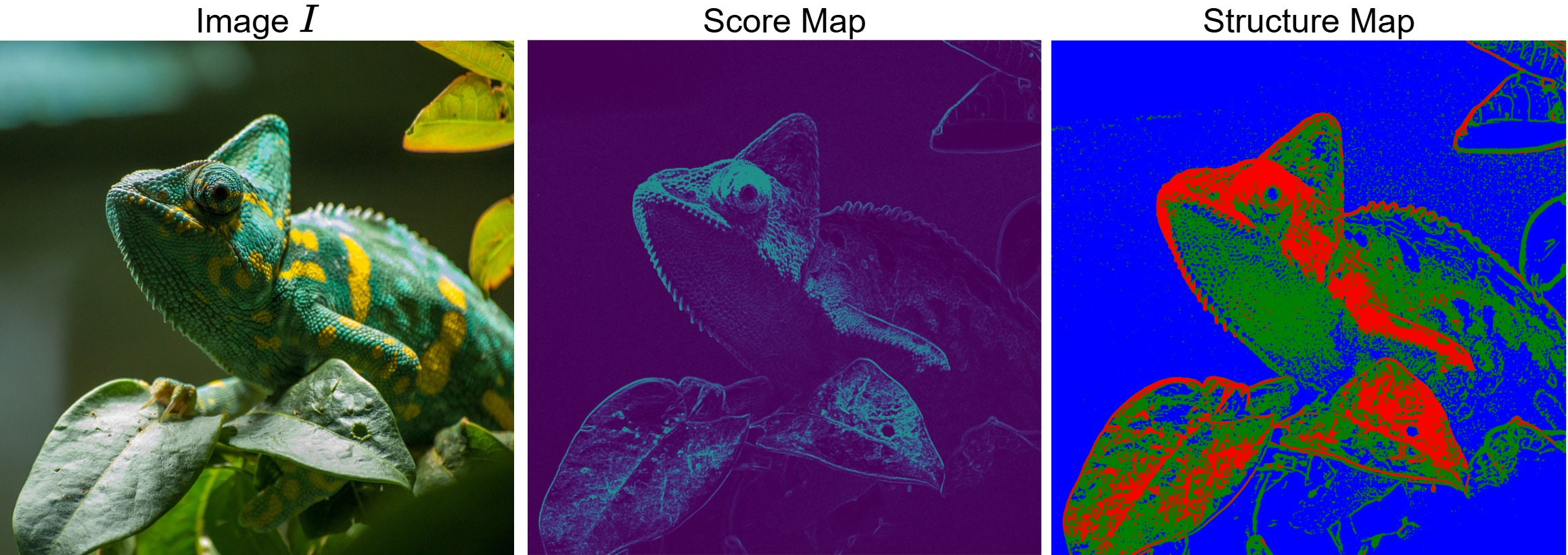}
\caption{\textbf{Structure Map construction:} Given an image, we compute a per-pixel score to generate the Score Map. Quantizing this map using two thresholds ($\varphi_1<\varphi_2$) yields the Structure Map.}
\label{fig:score_and_structure_map_example}
\end{figure}

We quantify the detail a region needs with a lightweight, parameter-free \textit{structure tensor} (gradient magnitude and local variance perform comparably, Tab.~\ref{tab:combined_ablation}). For an image $I\in\mathbb{R}^{3\times H'\times W'}$, we compute at each pixel the $2{\times}2$ second-moment matrix $\mathbf{S} = \left[\begin{smallmatrix} \sum I_x^2 & \sum I_x I_y \\ \sum I_x I_y & \sum I_y^2 \end{smallmatrix}\right]$ from local gradients $I_x, I_y$. Its eigenvalues $\lambda_1\ge\lambda_2\ge0$ characterize local content (near-zero: flat; one dominant: edge; two comparable: texture) combined into a per-pixel score
\begin{equation}
    \text{score} = \max(0,\, w_{texture}\lambda_2 + w_{edge}(\lambda_1-\lambda_2)).
\end{equation}
Assembling all pixel scores gives the \textit{Score Map}$\in\mathbb{R}^{H'\times W'}$, thresholded at $\varphi_1<\varphi_2$ into the \textbf{Structure Map}$\in\mathbb{Z}^{H'\times W'}$: level-0 ($\text{score}\le\varphi_1$, $\times1$), level-1 ($\varphi_1<\text{score}\le\varphi_2$, $\times2$), level-2 ($\text{score}>\varphi_2$, $\times4$) (Fig.~\ref{fig:score_and_structure_map_example}).

We compute this Score Map on the ground-truth $I_{SR}$, resize it to $\mathbf{F}_{base}$'s resolution, and quantize it into the training target. Since $I_{SR}$ is unavailable at inference, a lightweight Predictor ($\mathcal{L}_2$ loss) regresses the Score Map directly from $\mathbf{F}_{base}$.

\subsection{Sparse Feature Refinement}

\label{sec:hpc}




\subsubsection{Blind Feature Expansion}
\begin{figure}[H]
\centering
\includegraphics[width=0.9\columnwidth]{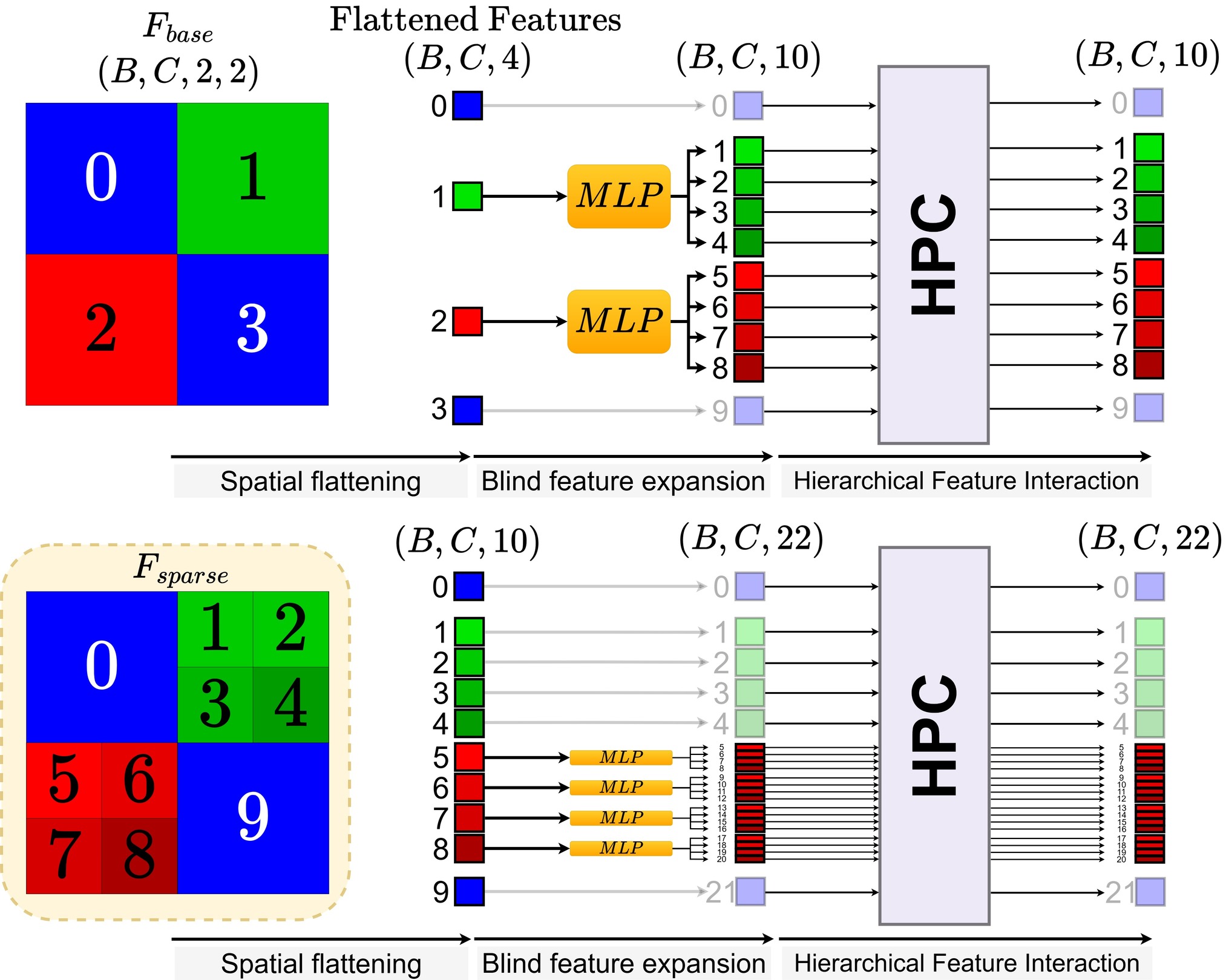}
\caption{\textbf{Two-stage blind expansion} of each feature, with \textbf{Hierarchical Pointer Convolution} applied after each stage to restore communication among topological neighbors.}

\label{fig:blind_exp}
\end{figure}

After the Structure Map assigns each feature in $\mathbf{F}_{base}$ its expansion factor ($\times1$, $\times2$ or $\times4$), we flatten $\mathbf{F}_{base}$ to $(B,C,H{\cdot}W)$. Features assigned $\times1$ (blue in Fig.~\ref{fig:blind_exp}) represent flat regions and need no further refinement, so we set them aside. The remaining $N_1$ features undergo a shared, initial $\times2$ expansion, where a lightweight MLP splits each feature into $4$ children. Among the $N_1$ parent features, those originally assigned $\times2$ are now complete and set aside, while those assigned $\times4$ undergo one more $\times2$ expansion. Since each feature expands independently, the resulting children are blind to one another; Sec.~\ref{sec:hfi} restores this missing communication before decoding into Gaussians.

\subsubsection{Hierarchical Feature Interaction}
\label{sec:hfi}
We restore this communication with a Hierarchical Pointer Convolution (HCP): for each feature, a convolution with its topological neighbors, retrieved via an index we call the Pointer Grid $\mathbf{P}$.

\begin{figure*}[t]
\centering
\includegraphics[width=0.97\textwidth]{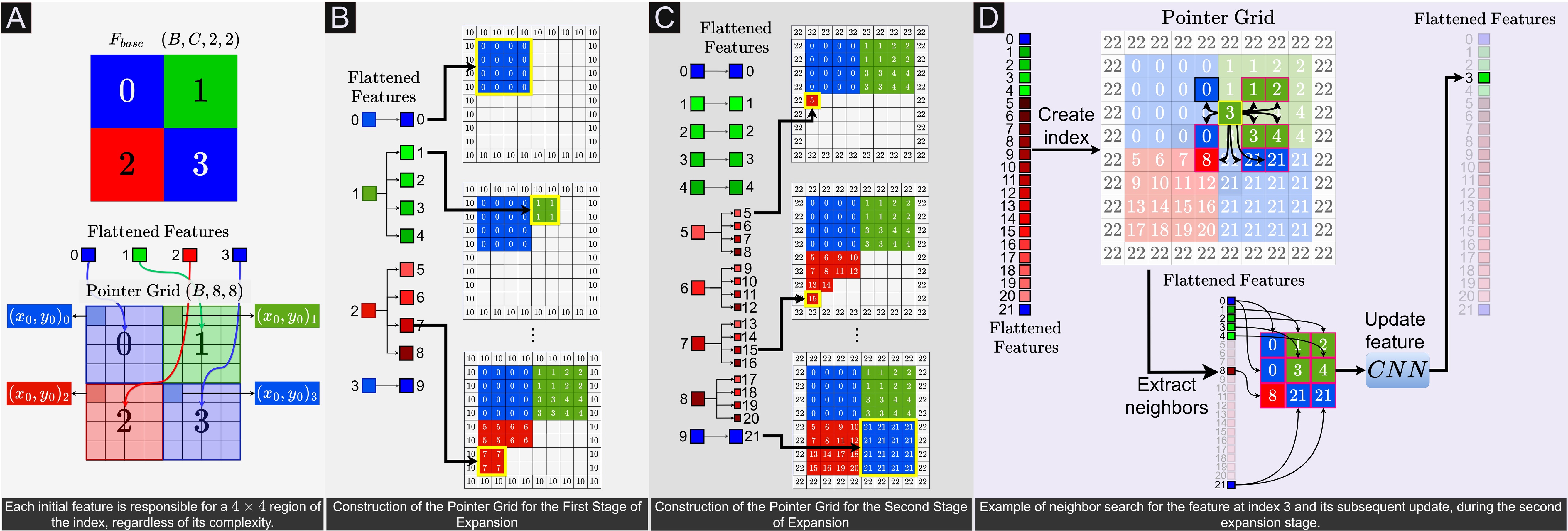}
\caption{\textbf{Pointer Grid construction and use.} After each expansion, sparse quadtree features are indexed by the Pointer Grid (A: roots, B: first expansion, C: second), enabling $O(1)$ retrieval of 9 neighbors for convolution. This defines \textbf{Hierarchical Pointer Convolution} (D).}

\label{fig:pointer_grid_and_retrieval}
\end{figure*}


\paragraph{Why not a generic sparse index?}
Retrieving an incorrect neighbor feeds the wrong feature into Gaussian blending, severely degrading reconstruction quality (Sec.~\ref{sec:ablation}). Generic sparse indices are designed for unpredictable data (e.g., point clouds) and fail in our single-pass setup for three reasons:
\textbf{1) Hash tables} (e.g., Instant-NGP) suffer from collisions. While optimization-based methods correct these over thousands of iterations, our single feed-forward pass cannot, turning collisions into permanent artifacts.
\textbf{2) 1D orderings} (e.g., Z-curves) destroy true 2D spatial adjacency, placing adjacent regions far apart.
\textbf{3) Tree traversals} require a variable number of steps per feature, stalling parallel GPU threads.
Conversely, our quadtree splits are strictly predictable. Instead of searching, we compute each child's address in closed form and write its ID directly into our dense 2D index (the Pointer Grid). This guarantees $O(1)$ spatial lookups with zero collisions, undistorted 2D topology, and no GPU divergence, by construction.

\paragraph{Building the index.}
The \textbf{Pointer Grid } $\mathbf{P}\in\mathbb{N}^{4H\times4W}$ maps the sparse quadtree topology onto a dense grid, storing the index of the feature responsible for each spatial location. Flattening the base feature map $\mathbf{F}_{base}$ from $(B,C,H,W)$ to $(B,C,H{\cdot}W)$ assigns to the feature at row $i$, column $j$ a flat index $n=iW{+}j$. This base feature sits at depth $d=0$ and is initially responsible for a full region of side $S=4$ with top-left corner $(4j,4i)$ (Fig.~\ref{fig:pointer_grid_and_retrieval}, A).

When a feature at depth $d$, owning a region of side $S=4/2^d$ with top-left corner $(x_0,y_0)$, splits into 4 children, each child $(a,b)\in\{0,1\}^2$ advances to depth $d{+}1$ and owns the $\tfrac{S}{2}{\times}\tfrac{S}{2}$ sub-region with top-left corner $(x_0,y_0)+\tfrac{S}{2}(a,b)$.

During the first expansion (Fig. \ref{fig:pointer_grid_and_retrieval}, B), roots assigned an expansion factor of $\times2$ or $\times4$ split to depth $d{=}1$ (owning $S{=}2$ regions), while flat ($\times1$) roots remain at $d{=}0$ (keeping $S{=}4$). This gives $M_1\ge HW$ features in total. Each feature writes its unique index $m$ into $\mathbf{P}$, covering its owned region in parallel:
\begin{equation}
    \mathbf{P}[y_0{:}y_0{+}S,\ x_0{:}x_0{+}S] = m.
    \label{eq:fill_pointer}
\end{equation}
The second expansion (Fig. \ref{fig:pointer_grid_and_retrieval}, C) repeats this process: only $d{=}1$ children flagged $\times4$ advance to $d{=}2$ (yielding $S{=}1$ regions), while the rest are left as is. This gives $M_2\ge M_1$ features, and again, each of them writes its own index in $\mathbf{P}$ using the same equation (\ref{eq:fill_pointer}).

\paragraph{Neighbor lookup and feature update.}
Because the quadtree leaves features at different depths and spatial resolutions, standard convolution offsets fail. Thanks to our dense Pointer Grid $\mathbf{P}$, we can instead efficiently sample a $3{\times}3$ neighborhood for any feature $m$ (owning a region of side $S$ at $(x_0,y_0)$) by defining 9 lookup coordinates. We compute the spatial offsets along the $y$ and $x$ axes as
\begin{equation}
    \mathcal{O}(S) = \{-1,\ \lfloor S/2 \rfloor,\ S\},
\end{equation}
and retrieve the 9 neighbor indices directly from $\mathbf{P}$:
\begin{equation}
    \mathcal{N}_m = \{\mathbf{P}[y_0{+}dy,\ x_0{+}dx] : dy,dx\in\mathcal{O}(S)\}.
\end{equation}


Using $\mathcal{N}_m$, we retrieve feature $m$'s 9 neighbors into a $C{\times}3{\times}3$ tensor and apply a standard convolution to compute its residual update. We call this step \textbf{Hierarchical Pointer Convolution (HPC)} (Fig.~\ref{fig:pointer_grid_and_retrieval}, D). Since this $O(1)$ sampling inherently approximates the true neighborhood and reaches only immediate neighbors, we stack multiple HPC blocks to effectively expand the receptive field.

\begin{figure*}[t]
\centering
\includegraphics[width=0.98\textwidth]{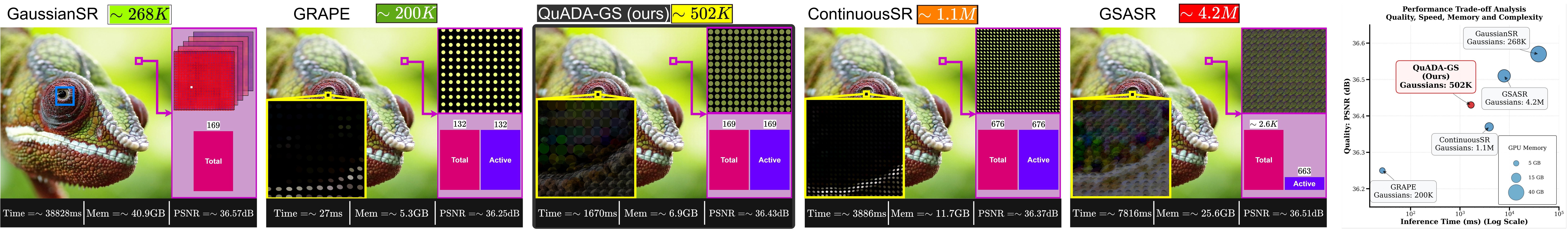}

\caption{\textbf{Performance trade-off:} GS-based ASR methods at $\times4$ SR (from $512{\times}512 \to 2048{\times}2048$). Total Gaussian count is shown next to each method name; zoomed patches detail Gaussian shapes and activations, with Time, Memory and PSNR at the bottom.}

\label{fig:explicit_comparison}
\end{figure*}

\subsection{Gaussian Decoding and Rasterization}
\label{sec:rasterization}

Each of the $M_2$ final features owns a region of size $S \in \{1, 2, 4\}$ at $(x_0, y_0)$ in $\mathbf{P}$. An MLP decodes them into continuous 2D Gaussians, predicting: opacity $\alpha \in [0,1]$, color $c \in [0,1]^3$, correlation $\rho \in [-1,1]$, offsets $(\Delta_x, \Delta_y) \in [-1/(4W), 1/(4H)]$, and scales $\sigma_x, \sigma_y \in (0,1)$. Multiplying predicted scales by $S$ ensures finer features ($S=1$) yield compact Gaussians while coarser ones ($S=4$) cover broader areas. The center is the region's midpoint plus the offset: $(\mu_x, \mu_y) = \big(\frac{x_0 + S/2}{4} + \Delta_x, \frac{y_0 + S/2}{4} + \Delta_y\big)$.

Following~\cite{Chen_2025_ICCV}, a Gaussian's color contribution at $(x,y)$ is $G(x,y) = \alpha \, c \, f(x,y)$, where $f$ is the anisotropic 2D Gaussian density parameterized by $\mu, \sigma, \rho$. The final super-resolved image $I_{SR}$ at target scale $s$ is rendered by summing all $N$ Gaussians evaluated at $(x/s, y/s)$:
\begin{equation}
    I_{SR}(x,y; s) = \sum_{i=1}^N G_i\big(x/s, y/s\big).
\end{equation}

\section{Experiments}
\label{sec:experiments}






\subsection{Implementation Details}

\paragraph{Network Architecture.} Following GSASR's Scale-Conditioned Feature Map Prediction~\cite{Chen_2025_ICCV}, we use EDSR-baseline~\cite{Lim_2017_CVPR_Workshops} or RDN~\cite{zhang2018residual} as feature extractor to obtain $\mathbf{F}_{LR} \in \mathbb{R}^{D \times H \times W}$ ($D=64$). We set base feature dimension $C=180$, Gaussian embedding grid $\mathbf{E}_{gs}$ size $K=12$, and Transformer refiner window size $12$. The W-MCA module has $2$ cross- and $6$ self-attention layers, followed by SW-MSA module with $12$ self-attention layers. The Pixel-Wise Upsampling Factor predictor (Sec.~\ref{sec:structure_map}) applies a lightweight CNN head (Conv2D-ReLU-Conv2D, $C \to C/2 \to 1$) to $\mathbf{F}_{base}$, producing a Score Map quantized via thresholds $\varphi_1=0.1, \varphi_2=1.4$ into the Structure Map. For the Hierarchical Feature Interaction (Sec.~\ref{sec:hfi}), we apply one Hierarchical Pointer Convolution block after the first expansion stage and two after the second. Gaussian parameters (Sec.~\ref{sec:rasterization}) are decoded via 3-layer MLPs with ReLU activations; predicted scales are multiplied by the spatial size $S \in \{1, 2, 4\}$ of their Pointer Grid region.

\paragraph{Training Details.} We train end-to-end with an $\mathcal{L}_1$ reconstruction loss between the predicted $\mathbf{I}_{SR}$ and the ground truth, plus an $\mathcal{L}_2$ loss (weight $0.1$) supervising the Pixel-Wise Upsampling Factor predictor, on the DIV2K dataset~\cite{Timofte_2017_CVPR_Workshops} using $48 \times 48$ LR patches and continuous random scales $s \sim \mathcal{U}(1, 4)$, strictly following competitor protocols. To rasterize, we employ the same CUDA rasterizer of GSASR ~\cite{Chen_2025_ICCV}. The model is optimized using Adam for $500$k iterations (initial learning rate $10^{-4}$, halved every $100$k steps) across three NVIDIA H100 GPUs.

\begin{table*}
\centering
\resizebox{\linewidth}{!}{
\setlength{\tabcolsep}{5pt}
\begin{tabular}{l rrrrrr rrrrrr rrrrrr rrrrrr}
\toprule
\multirow{2}{*}{Methods}
& \multicolumn{6}{c}{PSNR $\uparrow$}
& \multicolumn{6}{c}{SSIM $\uparrow$}
& \multicolumn{6}{c}{LPIPS $\downarrow$}
& \multicolumn{6}{c}{DISTS $\downarrow$} \\
\cmidrule(lr){2-7}
\cmidrule(lr){8-13}
\cmidrule(lr){14-19}
\cmidrule(lr){20-25}
& \multicolumn{1}{c}{$\times 2$}
& \multicolumn{1}{c}{$\times 3$}
& \multicolumn{1}{c}{$\times 4$}
& \multicolumn{1}{c}{$\times 6$}
& \multicolumn{1}{c}{$\times 8$}
& \multicolumn{1}{c}{$\times 12$}
& \multicolumn{1}{c}{$\times 2$}
& \multicolumn{1}{c}{$\times 3$}
& \multicolumn{1}{c}{$\times 4$}
& \multicolumn{1}{c}{$\times 6$}
& \multicolumn{1}{c}{$\times 8$}
& \multicolumn{1}{c}{$\times 12$}
& \multicolumn{1}{c}{$\times 2$}
& \multicolumn{1}{c}{$\times 3$}
& \multicolumn{1}{c}{$\times 4$}
& \multicolumn{1}{c}{$\times 6$}
& \multicolumn{1}{c}{$\times 8$}
& \multicolumn{1}{c}{$\times 12$}
& \multicolumn{1}{c}{$\times 2$}
& \multicolumn{1}{c}{$\times 3$}
& \multicolumn{1}{c}{$\times 4$}
& \multicolumn{1}{c}{$\times 6$}
& \multicolumn{1}{c}{$\times 8$}
& \multicolumn{1}{c}{$\times 12$} \\
\midrule
Meta-SR
& \RankCellEleven{32.05}{10} & \RankCellTwelve{28.10}{11} & \RankCellTwelve{25.94}{11} & \RankCellTwelve{23.57}{11} & \RankCellTwelve{22.27}{11} & \RankCellTwelve{20.77}{10}
& \RankCellEleven{0.9280}{10} & \RankCellTwelve{0.8520}{11} & \RankCellTwelve{0.7824}{11} & \RankCellTwelve{0.6726}{11} & \RankCellTwelve{0.6004}{11} & \RankCellTwelve{0.5191}{11}
& \RankCellEleven{0.0659}{9} & \RankCellTwelve{0.1591}{12} & \RankCellTwelve{0.2367}{12} & \RankCellTwelve{0.3540}{12} & \RankCellTwelve{0.4536}{12} & \RankCellTwelve{0.5818}{10}
& \RankCellEleven{0.0712}{7} & \RankCellTwelve{0.1277}{7} & \RankCellTwelve{0.1683}{6} & \RankCellTwelve{0.2312}{7} & \RankCellTwelve{0.2765}{9} & \RankCellTwelve{0.3397}{8} \\
LIIF
& \RankCellEleven{32.12}{8} & \RankCellTwelve{28.20}{10} & \RankCellTwelve{26.14}{10} & \RankCellTwelve{23.78}{9} & \RankCellTwelve{22.45}{8} & \RankCellTwelve{20.89}{8}
& \RankCellEleven{0.9288}{8} & \RankCellTwelve{0.8544}{9} & \RankCellTwelve{0.7885}{9} & \RankCellTwelve{0.6850}{8} & \RankCellTwelve{0.6169}{8} & \RankCellTwelve{0.5370}{8}
& \RankCellEleven{0.0643}{8} & \RankCellTwelve{0.1542}{10} & \RankCellTwelve{0.2271}{8} & \RankCellTwelve{0.3355}{8} & \RankCellTwelve{0.4201}{6} & \RankCellTwelve{0.5540}{6}
& \RankCellEleven{0.0717}{9} & \RankCellTwelve{0.1290}{10} & \RankCellTwelve{0.1738}{11} & \RankCellTwelve{0.2354}{10} & \RankCellTwelve{0.2795}{10} & \RankCellTwelve{0.3421}{10} \\
LTE
& \RankCellEleven{32.26}{6} & \RankCellTwelve{28.31}{7} & \RankCellTwelve{26.24}{7} & \RankCellTwelve{23.84}{7} & \RankCellTwelve{22.53}{7} & \RankCellTwelve{20.96}{7}
& \RankCellEleven{0.9299}{6} & \RankCellTwelve{0.8561}{7} & \RankCellTwelve{0.7910}{7} & \RankCellTwelve{0.6872}{7} & \RankCellTwelve{0.6190}{7} & \RankCellTwelve{0.5383}{7}
& \RankCellEleven{0.0629}{7} & \RankCellTwelve{0.1513}{7} & \RankCellTwelve{0.2223}{7} & \RankCellTwelve{0.3431}{10} & \RankCellTwelve{0.4341}{10} & \RankCellTwelve{0.5729}{9}
& \RankCellEleven{0.0711}{6} & \RankCellTwelve{0.1278}{8} & \RankCellTwelve{0.1718}{8} & \RankCellTwelve{0.2320}{8} & \RankCellTwelve{0.2763}{8} & \RankCellTwelve{0.3399}{9} \\
SRNO
& \RankCellEleven{32.56}{4} & \RankCellTwelve{28.54}{5} & \RankCellTwelve{26.48}{6} & \RankCellTwelve{24.07}{6} & \RankCellTwelve{22.69}{6} & \RankCellTwelve{21.10}{6}
& \RankCellEleven{0.9324}{4} & \RankCellTwelve{0.8599}{6} & \RankCellTwelve{0.7976}{6} & \RankCellTwelve{0.6956}{6} & \RankCellTwelve{0.6269}{6} & \RankCellTwelve{0.5454}{6}
& \RankCellEleven{0.0597}{5} & \RankCellTwelve{0.1465}{6} & \RankCellTwelve{0.2136}{5} & \RankCellTwelve{0.3292}{6} & \RankCellTwelve{0.4168}{5} & \RankCellTwelve{0.5544}{7}
& \RankCellEleven{0.0692}{4} & \RankCellTwelve{0.1243}{5} & \RankCellTwelve{0.1679}{5} & \RankCellTwelve{0.2258}{5} & \RankCellTwelve{0.2698}{5} & \RankCellTwelve{0.3333}{5} \\
LINF
& \RankCellEleven{32.12}{8} & \RankCellTwelve{28.21}{9} & \RankCellTwelve{26.16}{9} & \RankCellTwelve{23.79}{8} & \RankCellTwelve{22.45}{8} & \RankCellTwelve{20.88}{9}
& \RankCellEleven{0.9283}{9} & \RankCellTwelve{0.8538}{10} & \RankCellTwelve{0.7878}{10} & \RankCellTwelve{0.6836}{9} & \RankCellTwelve{0.6141}{9} & \RankCellTwelve{0.5337}{9}
& \RankCellEleven{0.0660}{10} & \RankCellTwelve{0.1570}{11} & \RankCellTwelve{0.2308}{10} & \RankCellTwelve{0.3425}{9} & \RankCellTwelve{0.4301}{8} & \RankCellTwelve{0.5676}{8}
& \RankCellEleven{0.0717}{9} & \RankCellTwelve{0.1291}{11} & \RankCellTwelve{0.1748}{12} & \RankCellTwelve{0.2374}{12} & \RankCellTwelve{0.2831}{11} & \RankCellTwelve{0.3479}{11} \\
LMF
& \RankCellEleven{32.48}{5} & \RankCellTwelve{28.59}{4} & \RankCellTwelve{26.50}{4} & \RankCellTwelve{24.08}{5} & \RankCellTwelve{22.73}{5} & \RankCellTwelve{21.12}{5}
& \RankCellEleven{0.9323}{5} & \RankCellTwelve{0.8615}{5} & \RankCellTwelve{0.7985}{5} & \RankCellTwelve{0.6968}{5} & \RankCellTwelve{0.6286}{5} & \RankCellTwelve{0.5460}{5}
& \RankCellEleven{0.0603}{6} & \RankCellTwelve{0.1437}{5} & \RankCellTwelve{0.2151}{6} & \RankCellTwelve{0.3327}{7} & \RankCellTwelve{0.4205}{7} & \RankCellTwelve{0.5405}{5}
& \RankCellEleven{0.0696}{5} & \RankCellTwelve{0.1250}{6} & \RankCellTwelve{0.1696}{7} & \RankCellTwelve{0.2300}{6} & \RankCellTwelve{0.2748}{6} & \RankCellTwelve{0.3388}{7} \\
Ciao-SR
& \RankCellEleven{32.79}{3} & \RankCellTwelve{28.67}{3} & \RankCellTwelve{26.69}{3} & \RankCellTwelve{24.23}{4} & \RankCellTwelve{22.83}{4} & \RankCellTwelve{21.19}{4}
& \RankCellEleven{0.9344}{3} & \RankCellTwelve{0.8627}{4} & \RankCellTwelve{0.8031}{4} & \RankCellTwelve{0.7029}{4} & \RankCellTwelve{0.6344}{4} & \RankCellTwelve{0.5530}{4}
& \RankCellEleven{0.0580}{3} & \RankCellTwelve{0.1408}{4} & \RankCellTwelve{0.2078}{3} & \RankCellTwelve{0.3117}{4} & \RankCellTwelve{0.3932}{3} & \RankCellTwelve{0.5204}{3}
& \RankCellEleven{0.0676}{3} & \RankCellTwelve{0.1181}{4} & \RankCellTwelve{0.1659}{4} & \RankCellTwelve{0.2250}{4} & \RankCellTwelve{0.2666}{4} & \RankCellTwelve{0.3278}{4} \\
\midrule
Gaussian-SR
& \RankCellEleven{32.22}{7} & \RankCellTwelve{28.27}{8} & \RankCellTwelve{26.19}{8} & \RankCellTwelve{23.77}{10} & \RankCellTwelve{22.36}{10} & \RankCellTwelve{20.68}{12}
& \RankCellEleven{0.9296}{7} & \RankCellTwelve{0.8553}{8} & \RankCellTwelve{0.7893}{8} & \RankCellTwelve{0.6814}{10} & \RankCellTwelve{0.6070}{10} & \RankCellTwelve{0.5217}{10}
& \RankCellEleven{0.0590}{4} & \RankCellTwelve{0.1541}{8} & \RankCellTwelve{0.2283}{9} & \RankCellTwelve{0.3499}{11} & \RankCellTwelve{0.4530}{11} & \RankCellTwelve{0.6198}{12}
& \RankCellEleven{0.0712}{7} & \RankCellTwelve{0.1288}{9} & \RankCellTwelve{0.1730}{10} & \RankCellTwelve{0.2364}{11} & \RankCellTwelve{0.2883}{12} & \RankCellTwelve{0.3646}{12} \\
CSR*
& - & \RankCellTwelve{28.48}{6} & \RankCellTwelve{26.50}{4} & \RankCellTwelve{24.25}{3} & \RankCellTwelve{22.95}{3} & \RankCellTwelve{21.41}{2}
& - & \RankCellTwelve{0.8683}{3} & \RankCellTwelve{0.8104}{3} & \RankCellTwelve{0.7183}{2} & \RankCellTwelve{0.6505}{2} & \RankCellTwelve{0.5662}{2}
& - & \RankCellTwelve{0.1356}{3} & \RankCellTwelve{0.2080}{4} & \RankCellTwelve{0.3037}{2} & \RankCellTwelve{0.3982}{4} & \RankCellTwelve{0.5236}{4}
& - & \RankCellTwelve[\textbf{0.1149}]{0.1149}{1} & \RankCellTwelve[\textbf{0.1541}]{0.1541}{1} & \RankCellTwelve[\textbf{0.2098}]{0.2098}{1} & \RankCellTwelve[\textbf{0.2487}]{0.2487}{1} & \RankCellTwelve[\textbf{0.3084}]{0.3084}{1} \\
GRAPE
& \RankCellEleven{30.08}{11} & \RankCellTwelve{27.61}{12} & \RankCellTwelve{25.86}{12} & \RankCellTwelve{23.51}{12} & \RankCellTwelve{22.24}{12} & \RankCellTwelve{20.74}{11}
& \RankCellEleven{0.9169}{11} & \RankCellTwelve{0.8365}{12} & \RankCellTwelve{0.7806}{12} & \RankCellTwelve{0.6659}{12} & \RankCellTwelve{0.5970}{12} & \RankCellTwelve{0.5185}{12}
& \RankCellEleven{0.0667}{11} & \RankCellTwelve{0.1541}{8} & \RankCellTwelve{0.2322}{11} & \RankCellTwelve{0.3283}{5} & \RankCellTwelve{0.4327}{9} & \RankCellTwelve{0.5848}{11}
& \RankCellEleven{0.0759}{11} & \RankCellTwelve{0.1376}{12} & \RankCellTwelve{0.1721}{9} & \RankCellTwelve{0.2340}{9} & \RankCellTwelve{0.2759}{7} & \RankCellTwelve{0.3380}{6} \\
GSASR
& \RankCellEleven[\textbf{33.27}]{33.27}{1} & \RankCellTwelve{29.17}{2} & \RankCellTwelve{27.01}{2} & \RankCellTwelve{24.51}{2} & \RankCellTwelve{23.09}{2} & \RankCellTwelve{21.39}{3}
& \RankCellEleven{0.9389}{2} & \RankCellTwelve{0.8733}{2} & \RankCellTwelve{0.8142}{2} & \RankCellTwelve{0.7165}{3} & \RankCellTwelve{0.6480}{3} & \RankCellTwelve{0.5628}{3}
& \RankCellEleven[\textbf{0.0525}]{0.0525}{1} & \RankCellTwelve[\textbf{0.1320}]{0.1320}{1} & \RankCellTwelve[\textbf{0.1987}]{0.1987}{1} & \RankCellTwelve[\textbf{0.2975}]{0.2975}{1} & \RankCellTwelve[\textbf{0.3754}]{0.3754}{1} & \RankCellTwelve[\textbf{0.4994}]{0.4994}{1}
& \RankCellEleven[\textbf{0.0634}]{0.0634}{1} & \RankCellTwelve{0.1156}{2} & \RankCellTwelve{0.1552}{2} & \RankCellTwelve{0.2111}{2} & \RankCellTwelve{0.2515}{2} & \RankCellTwelve{0.3135}{2} \\
\textbf{QuADA-GS (ours)}
& \RankCellEleven{33.20}{2} & \RankCellTwelve[\textbf{29.38}]{29.38}{1} & \RankCellTwelve[\textbf{27.23}]{27.23}{1} & \RankCellTwelve[\textbf{24.70}]{24.70}{1} & \RankCellTwelve[\textbf{23.25}]{23.25}{1} & \RankCellTwelve[\textbf{21.50}]{21.50}{1}
& \RankCellEleven[\textbf{0.9393}]{0.9393}{1} & \RankCellTwelve[\textbf{0.8762}]{0.8762}{1} & \RankCellTwelve[\textbf{0.8188}]{0.8188}{1} & \RankCellTwelve[\textbf{0.7234}]{0.7234}{1} & \RankCellTwelve[\textbf{0.6559}]{0.6559}{1} & \RankCellTwelve[\textbf{0.5716}]{0.5716}{1}
& \RankCellEleven{0.0535}{2} & \RankCellTwelve{0.1330}{2} & \RankCellTwelve{0.2017}{2} & \RankCellTwelve{0.3048}{3} & \RankCellTwelve{0.3853}{2} & \RankCellTwelve{0.5116}{2}
& \RankCellEleven{0.0654}{2} & \RankCellTwelve{0.1159}{3} & \RankCellTwelve{0.1557}{3} & \RankCellTwelve{0.2136}{3} & \RankCellTwelve{0.2571}{3} & \RankCellTwelve{0.3197}{3} \\
\midrule
\textbf{QuADA-GS (ULTRA) (ours)} & \textcolor{green!60!black}{\textbf{33.64}} & \textcolor{green!60!black}{\textbf{29.57}} & \textcolor{green!60!black}{\textbf{27.42}} & \textcolor{green!60!black}{\textbf{24.86}} & \textcolor{green!60!black}{\textbf{23.39}} & \textcolor{green!60!black}{\textbf{21.57}} & \textcolor{green!60!black}{\textbf{0.9413}} & \textcolor{green!60!black}{\textbf{0.8785}} & \textcolor{green!60!black}{\textbf{0.8224}} & \textcolor{green!60!black}{\textbf{0.7293}} & \textcolor{green!60!black}{\textbf{0.6627}} & \textcolor{green!60!black}{\textbf{0.5774}} & \textcolor{green!60!black}{\textbf{0.0481}} & \textcolor{green!60!black}{\textbf{0.1275}} & \textcolor{green!60!black}{\textbf{0.1921}} & \textcolor{green!60!black}{\textbf{0.2887}} & \textcolor{green!60!black}{\textbf{0.3643}} & \textcolor{green!60!black}{\textbf{0.4865}} & \textcolor{green!60!black}{\textbf{0.0621}} & 0.1156 & 0.1557 & 0.2127 & 0.2528 & 0.3125 \\
\bottomrule
\end{tabular}
}
\end{table*}

\begin{table*}
\centering
\resizebox{\linewidth}{!}{
\setlength{\tabcolsep}{5pt}
\begin{tabular}{l rrrrrr rrrrrr rrrrrr rrrrrr}
\toprule
\multirow{2}{*}{Methods}
& \multicolumn{6}{c}{PSNR $\uparrow$}
& \multicolumn{6}{c}{SSIM $\uparrow$}
& \multicolumn{6}{c}{LPIPS $\downarrow$}
& \multicolumn{6}{c}{DISTS $\downarrow$} \\
\cmidrule(lr){2-7}
\cmidrule(lr){8-13}
\cmidrule(lr){14-19}
\cmidrule(lr){20-25}
& \multicolumn{1}{c}{$\times 2$}
& \multicolumn{1}{c}{$\times 3$}
& \multicolumn{1}{c}{$\times 4$}
& \multicolumn{1}{c}{$\times 6$}
& \multicolumn{1}{c}{$\times 8$}
& \multicolumn{1}{c}{$\times 12$}
& \multicolumn{1}{c}{$\times 2$}
& \multicolumn{1}{c}{$\times 3$}
& \multicolumn{1}{c}{$\times 4$}
& \multicolumn{1}{c}{$\times 6$}
& \multicolumn{1}{c}{$\times 8$}
& \multicolumn{1}{c}{$\times 12$}
& \multicolumn{1}{c}{$\times 2$}
& \multicolumn{1}{c}{$\times 3$}
& \multicolumn{1}{c}{$\times 4$}
& \multicolumn{1}{c}{$\times 6$}
& \multicolumn{1}{c}{$\times 8$}
& \multicolumn{1}{c}{$\times 12$}
& \multicolumn{1}{c}{$\times 2$}
& \multicolumn{1}{c}{$\times 3$}
& \multicolumn{1}{c}{$\times 4$}
& \multicolumn{1}{c}{$\times 6$}
& \multicolumn{1}{c}{$\times 8$}
& \multicolumn{1}{c}{$\times 12$} \\
\midrule
Meta-SR
& \RankCellEleven{33.55}{10} & \RankCellTwelve{30.27}{10} & \RankCellTwelve{28.51}{10} & \RankCellTwelve{26.28}{11} & \RankCellTwelve{24.75}{12} & \RankCellTwelve{23.05}{9}
& \RankCellEleven{0.9177}{10} & \RankCellTwelve{0.8421}{11} & \RankCellTwelve{0.7808}{11} & \RankCellTwelve{0.6897}{11} & \RankCellTwelve{0.6300}{11} & \RankCellTwelve{0.5616}{11}
& \RankCellEleven{0.0953}{10} & \RankCellTwelve{0.2110}{12} & \RankCellTwelve{0.2882}{12} & \RankCellTwelve{0.3957}{11} & \RankCellTwelve{0.4674}{11} & \RankCellTwelve{0.5587}{9}
& \RankCellEleven{0.0845}{9} & \RankCellTwelve{0.1322}{11} & \RankCellTwelve{0.1636}{7} & \RankCellTwelve{0.2204}{11} & \RankCellTwelve{0.2578}{6} & \RankCellTwelve{0.3078}{4} \\
LIIF
& \RankCellEleven{33.63}{8} & \RankCellTwelve{30.32}{9} & \RankCellTwelve{28.60}{9} & \RankCellTwelve{26.44}{8} & \RankCellTwelve{24.92}{7} & \RankCellTwelve{23.15}{7}
& \RankCellEleven{0.9186}{8} & \RankCellTwelve{0.8431}{8} & \RankCellTwelve{0.7828}{8} & \RankCellTwelve{0.6964}{8} & \RankCellTwelve{0.6395}{8} & \RankCellTwelve{0.5738}{6}
& \RankCellEleven{0.0943}{9} & \RankCellTwelve{0.2067}{10} & \RankCellTwelve{0.2832}{8} & \RankCellTwelve{0.3861}{6} & \RankCellTwelve{0.4459}{5} & \RankCellTwelve{0.5465}{5}
& \RankCellEleven{0.0839}{6} & \RankCellTwelve{0.1311}{9} & \RankCellTwelve{0.1639}{10} & \RankCellTwelve{0.2197}{9} & \RankCellTwelve{0.2594}{10} & \RankCellTwelve{0.3117}{7} \\
LTE
& \RankCellEleven{33.68}{6} & \RankCellTwelve{30.35}{6} & \RankCellTwelve{28.63}{6} & \RankCellTwelve{26.48}{6} & \RankCellTwelve{24.96}{6} & \RankCellTwelve{23.20}{4}
& \RankCellEleven{0.9190}{6} & \RankCellTwelve{0.8437}{6} & \RankCellTwelve{0.7836}{6} & \RankCellTwelve{0.6972}{6} & \RankCellTwelve{0.6402}{7} & \RankCellTwelve{0.5741}{5}
& \RankCellEleven{0.0927}{5} & \RankCellTwelve{0.2045}{8} & \RankCellTwelve{0.2828}{7} & \RankCellTwelve{0.3919}{9} & \RankCellTwelve{0.4593}{9} & \RankCellTwelve{0.5638}{10}
& \RankCellEleven{0.0844}{8} & \RankCellTwelve{0.1320}{10} & \RankCellTwelve{0.1637}{8} & \RankCellTwelve{0.2193}{7} & \RankCellTwelve{0.2595}{11} & \RankCellTwelve{0.3162}{11} \\
SRNO
& \RankCellEleven{33.78}{4} & \RankCellTwelve{30.47}{3} & \RankCellTwelve{28.74}{4} & \RankCellTwelve{26.53}{4} & \RankCellTwelve{25.03}{4} & \RankCellTwelve{23.20}{4}
& \RankCellEleven{0.9201}{4} & \RankCellTwelve{0.8448}{5} & \RankCellTwelve{0.7856}{4} & \RankCellTwelve{0.6989}{4} & \RankCellTwelve{0.6456}{3} & \RankCellTwelve{0.5758}{4}
& \RankCellEleven{0.0920}{4} & \RankCellTwelve{0.2038}{7} & \RankCellTwelve{0.2790}{5} & \RankCellTwelve{0.3890}{7} & \RankCellTwelve{0.4515}{7} & \RankCellTwelve{0.5543}{8}
& \RankCellEleven{0.0830}{5} & \RankCellTwelve{0.1303}{8} & \RankCellTwelve{0.1638}{9} & \RankCellTwelve{0.2184}{6} & \RankCellTwelve{0.2583}{8} & \RankCellTwelve{0.3139}{9} \\
LINF
& \RankCellEleven{33.63}{8} & \RankCellTwelve{30.33}{8} & \RankCellTwelve{28.62}{7} & \RankCellTwelve{26.45}{7} & \RankCellTwelve{24.92}{7} & \RankCellTwelve{23.17}{6}
& \RankCellEleven{0.9178}{9} & \RankCellTwelve{0.8430}{9} & \RankCellTwelve{0.7826}{9} & \RankCellTwelve{0.6960}{9} & \RankCellTwelve{0.6386}{9} & \RankCellTwelve{0.5718}{8}
& \RankCellEleven{0.0970}{11} & \RankCellTwelve{0.2081}{11} & \RankCellTwelve{0.2861}{11} & \RankCellTwelve{0.3900}{8} & \RankCellTwelve{0.4526}{8} & \RankCellTwelve{0.5476}{6}
& \RankCellEleven{0.0841}{7} & \RankCellTwelve{0.1302}{7} & \RankCellTwelve{0.1642}{11} & \RankCellTwelve{0.2201}{10} & \RankCellTwelve{0.2589}{9} & \RankCellTwelve{0.3138}{8} \\
LMF
& \RankCellEleven{33.75}{5} & \RankCellTwelve{30.41}{5} & \RankCellTwelve{28.69}{5} & \RankCellTwelve{26.50}{5} & \RankCellTwelve{25.00}{5} & \RankCellTwelve{23.09}{8}
& \RankCellEleven{0.9196}{5} & \RankCellTwelve{0.8449}{4} & \RankCellTwelve{0.7849}{5} & \RankCellTwelve{0.6968}{7} & \RankCellTwelve{0.6403}{6} & \RankCellTwelve{0.5706}{9}
& \RankCellEleven{0.0916}{3} & \RankCellTwelve{0.2005}{5} & \RankCellTwelve{0.2794}{6} & \RankCellTwelve{0.3939}{10} & \RankCellTwelve{0.4595}{10} & \RankCellTwelve{0.5686}{11}
& \RankCellEleven{0.0829}{4} & \RankCellTwelve{0.1297}{6} & \RankCellTwelve{0.1618}{5} & \RankCellTwelve{0.2195}{8} & \RankCellTwelve{0.2582}{7} & \RankCellTwelve{0.3145}{10} \\
Ciao-SR
& \RankCellEleven{33.90}{3} & \RankCellTwelve{30.46}{4} & \RankCellTwelve{28.77}{3} & \RankCellTwelve{26.60}{3} & \RankCellTwelve{25.09}{3} & \RankCellTwelve[\textbf{23.27}]{23.27}{1}
& \RankCellEleven{0.9203}{3} & \RankCellTwelve{0.8450}{3} & \RankCellTwelve{0.7865}{3} & \RankCellTwelve{0.7003}{3} & \RankCellTwelve{0.6450}{4} & \RankCellTwelve{0.5788}{3}
& \RankCellEleven{0.0929}{6} & \RankCellTwelve{0.2018}{6} & \RankCellTwelve{0.2783}{3} & \RankCellTwelve{0.3823}{5} & \RankCellTwelve{0.4395}{3} & \RankCellTwelve{0.5340}{2}
& \RankCellEleven{0.0821}{3} & \RankCellTwelve{0.1271}{2} & \RankCellTwelve{0.1615}{4} & \RankCellTwelve{0.2163}{5} & \RankCellTwelve{0.2541}{5} & \RankCellTwelve{0.3091}{5} \\
\midrule
Gaussian-SR
& \RankCellEleven{33.64}{7} & \RankCellTwelve{30.34}{7} & \RankCellTwelve{28.62}{7} & \RankCellTwelve{26.41}{9} & \RankCellTwelve{24.80}{10} & \RankCellTwelve{22.92}{12}
& \RankCellEleven{0.9189}{7} & \RankCellTwelve{0.8435}{7} & \RankCellTwelve{0.7831}{7} & \RankCellTwelve{0.6936}{10} & \RankCellTwelve{0.6330}{10} & \RankCellTwelve{0.5621}{10}
& \RankCellEleven{0.0936}{8} & \RankCellTwelve{0.2056}{9} & \RankCellTwelve{0.2849}{9} & \RankCellTwelve{0.3973}{12} & \RankCellTwelve{0.4683}{12} & \RankCellTwelve{0.5903}{12}
& \RankCellEleven{0.0845}{9} & \RankCellTwelve{0.1327}{12} & \RankCellTwelve{0.1646}{12} & \RankCellTwelve{0.2221}{12} & \RankCellTwelve{0.2650}{12} & \RankCellTwelve{0.3254}{12} \\
CSR*
& - & \RankCellTwelve{30.19}{11} & \RankCellTwelve{28.37}{12} & \RankCellTwelve{26.29}{10} & \RankCellTwelve{24.84}{9} & \RankCellTwelve{22.98}{11}
& - & \RankCellTwelve{0.8423}{10} & \RankCellTwelve{0.7815}{10} & \RankCellTwelve{0.6975}{5} & \RankCellTwelve{0.6418}{5} & \RankCellTwelve{0.5737}{7}
& - & \RankCellTwelve{0.1986}{3} & \RankCellTwelve{0.2782}{2} & \RankCellTwelve{0.3725}{2} & \RankCellTwelve{0.4490}{6} & \RankCellTwelve{0.5461}{4}
& - & \RankCellTwelve{0.1271}{2} & \RankCellTwelve{0.1579}{2} & \RankCellTwelve[\textbf{0.2107}]{0.2107}{1} & \RankCellTwelve[\textbf{0.2463}]{0.2463}{1} & \RankCellTwelve[\textbf{0.3024}]{0.3024}{1} \\
GRAPE
& \RankCellEleven{32.21}{11} & \RankCellTwelve{29.87}{12} & \RankCellTwelve{28.45}{11} & \RankCellTwelve{26.23}{12} & \RankCellTwelve{24.80}{10} & \RankCellTwelve{23.02}{10}
& \RankCellEleven{0.9133}{11} & \RankCellTwelve{0.8306}{12} & \RankCellTwelve{0.7804}{12} & \RankCellTwelve{0.6836}{12} & \RankCellTwelve{0.6275}{12} & \RankCellTwelve{0.5599}{12}
& \RankCellEleven{0.0930}{7} & \RankCellTwelve[\textbf{0.1594}]{0.1594}{1} & \RankCellTwelve{0.2857}{10} & \RankCellTwelve[\textbf{0.3687}]{0.3687}{1} & \RankCellTwelve{0.4364}{2} & \RankCellTwelve{0.5530}{7}
& \RankCellEleven{0.0863}{11} & \RankCellTwelve{0.1272}{4} & \RankCellTwelve{0.1628}{6} & \RankCellTwelve{0.2137}{4} & \RankCellTwelve{0.2510}{4} & \RankCellTwelve{0.3108}{6} \\
GSASR
& \RankCellEleven{34.02}{2} & \RankCellTwelve[\textbf{30.65}]{30.65}{1} & \RankCellTwelve[\textbf{28.88}]{28.88}{1} & \RankCellTwelve[\textbf{26.71}]{26.71}{1} & \RankCellTwelve[\textbf{25.23}]{25.23}{1} & \RankCellTwelve{23.24}{3}
& \RankCellEleven{0.9222}{2} & \RankCellTwelve[\textbf{0.8498}]{0.8498}{1} & \RankCellTwelve[\textbf{0.7902}]{0.7902}{1} & \RankCellTwelve[\textbf{0.7058}]{0.7058}{1} & \RankCellTwelve[\textbf{0.6505}]{0.6505}{1} & \RankCellTwelve{0.5813}{2}
& \RankCellEleven[\textbf{0.0846}]{0.0846}{1} & \RankCellTwelve{0.1928}{2} & \RankCellTwelve[\textbf{0.2680}]{0.2680}{1} & \RankCellTwelve{0.3731}{3} & \RankCellTwelve[\textbf{0.4359}]{0.4359}{1} & \RankCellTwelve[\textbf{0.5306}]{0.5306}{1}
& \RankCellEleven[\textbf{0.0793}]{0.0793}{1} & \RankCellTwelve[\textbf{0.1267}]{0.1267}{1} & \RankCellTwelve[\textbf{0.1574}]{0.1574}{1} & \RankCellTwelve{0.2108}{2} & \RankCellTwelve{0.2480}{2} & \RankCellTwelve{0.3049}{3} \\
\textbf{QuADA-GS (ours)}
& \RankCellEleven[\textbf{34.03}]{34.03}{1} & \RankCellTwelve{30.55}{2} & \RankCellTwelve{28.80}{2} & \RankCellTwelve{26.62}{2} & \RankCellTwelve{25.20}{2} & \RankCellTwelve{23.26}{2}
& \RankCellEleven[\textbf{0.9232}]{0.9232}{1} & \RankCellTwelve{0.8487}{2} & \RankCellTwelve{0.7891}{2} & \RankCellTwelve{0.7035}{2} & \RankCellTwelve{0.6489}{2} & \RankCellTwelve[\textbf{0.5822}]{0.5822}{1}
& \RankCellEleven{0.0871}{2} & \RankCellTwelve{0.1998}{4} & \RankCellTwelve{0.2785}{4} & \RankCellTwelve{0.3795}{4} & \RankCellTwelve{0.4435}{4} & \RankCellTwelve{0.5410}{3}
& \RankCellEleven{0.0804}{2} & \RankCellTwelve{0.1284}{5} & \RankCellTwelve{0.1594}{3} & \RankCellTwelve{0.2132}{3} & \RankCellTwelve{0.2484}{3} & \RankCellTwelve{0.3045}{2} \\
\midrule
\textbf{QuADA-GS (ULTRA) (ours)} & \textcolor{green!60!black}{\textbf{34.15}} & \textcolor{green!60!black}{\textbf{30.68}} & \textcolor{green!60!black}{\textbf{28.91}} & \textcolor{green!60!black}{\textbf{26.74}} & \textcolor{green!60!black}{\textbf{25.26}} & \textcolor{green!60!black}{\textbf{23.28}} & \textcolor{green!60!black}{\textbf{0.9237}} & \textcolor{green!60!black}{\textbf{0.8501}} & \textcolor{green!60!black}{\textbf{0.7907}} & \textcolor{green!60!black}{\textbf{0.7067}} & \textcolor{green!60!black}{\textbf{0.6519}} & \textcolor{green!60!black}{\textbf{0.5853}} & \textcolor{green!60!black}{\textbf{0.0839}} & 0.1947 & 0.2691 & 0.3702 & \textcolor{green!60!black}{\textbf{0.4315}} & \textcolor{green!60!black}{\textbf{0.5244}} & 0.0796 & 0.1280 & 0.1589 & \textcolor{green!60!black}{\textbf{0.2107}} & 0.2472 & 0.3033 \\
\bottomrule
\end{tabular}
}
\caption{Quantitative results on Urban100 \cite{urban100} and Set14 \cite{set14}. All models use the EDSR-baseline \cite{Lim_2017_CVPR_Workshops} backbone and are evaluated across continuous scales ($\times 2$ to $\times 12$). PSNR/SSIM are computed on the YCbCr Y-channel. Best results in \textbf{bold}.}
\label{tab:urban_and_set14}
\end{table*}

\paragraph{Compared models.}
We evaluate against state-of-the-art \emph{implicit} architectures (Meta-SR~\cite{hu2019meta}, LIIF~\cite{chen2021learning}, LTE~\cite{lee2022local}, SRNO~\cite{Wei_2023_CVPR}, LINF~\cite{Yao_2023_CVPR}, CiaoSR~\cite{cao2023ciaosr}, LMF~\cite{He_2024_CVPR}) alongside recent \emph{explicit} splatting-based frameworks (ContinuousSR~\cite{continuoussr}, GRAPE~\cite{grape}, GaussianSR~\cite{hu2025gaussiansr}, GSASR~\cite{Chen_2025_ICCV}).

\paragraph{Evaluation protocol.}
We benchmark across standard datasets (Set5~\cite{set5}, Set14~\cite{set14}, DIV2K~\cite{Timofte_2017_CVPR_Workshops}, Urban100~\cite{urban100}, BSDS100~\cite{bsds100}, Manga109~\cite{manga109}, General100~\cite{GENERAL100}, LSDIR~\cite{lsdir_dataset}) for typical upsampling factors ($s \in \{2, 3, 4, 6, 8, 12\}$), deferring extreme out-of-distribution scales ($s \in \{16, 18, 24, 30\}$) on DIV2K and LSDIR to the Supplement. We compute PSNR/SSIM~\cite{SSIM} (Y-channel), LPIPS~\cite{lpips}, and DISTS~\cite{dists} using official scripts. Following GSASR~\cite{Chen_2025_ICCV}, hardware efficiency is measured end-to-end on a single H100 GPU using $720^2$ HR center crops from DIV2K. Note that for ContinuousSR (CSR), 
the released model uses HAT \cite{hat_backbone} as the feature extractor.

\subsection{Discussion}

\textbf{Fidelity–Efficiency Trade-off via Densification.} QuADA-GS achieves an optimal fidelity-efficiency balance by allocating Gaussian density strictly where needed. As Fig.~\ref{fig:explicit_comparison} highlights, GSASR requires over 4.2M Gaussians, over-allocating flat areas (45K allocated vs.\ 6.6K active) for a marginal $+0.08$~dB gain. QuADA-GS matches this fidelity using just $502$K Gaussians. By deploying fewer, larger Gaussians in flat regions and adaptively increasing density only to model complex details, it runs $5\times$ faster and $3.7\times$ lighter. It strictly dominates ContinuousSR ($2.3\times$ faster, half the memory, $+0.06$~dB) and vastly outperforms GaussianSR in efficiency ($23\times$ faster, $6\times$ lighter) for a minor $0.14$~dB trade-off. At $512\times512$, all methods, including lightweight ones like GRAPE, look visually similar, as the low-resolution input already retains sufficient high-frequency structure. But \textit{this example primarily highlights how computational costs scale for large inputs}.

\begin{figure*}[t]
\centering
\includegraphics[width=0.922\textwidth]{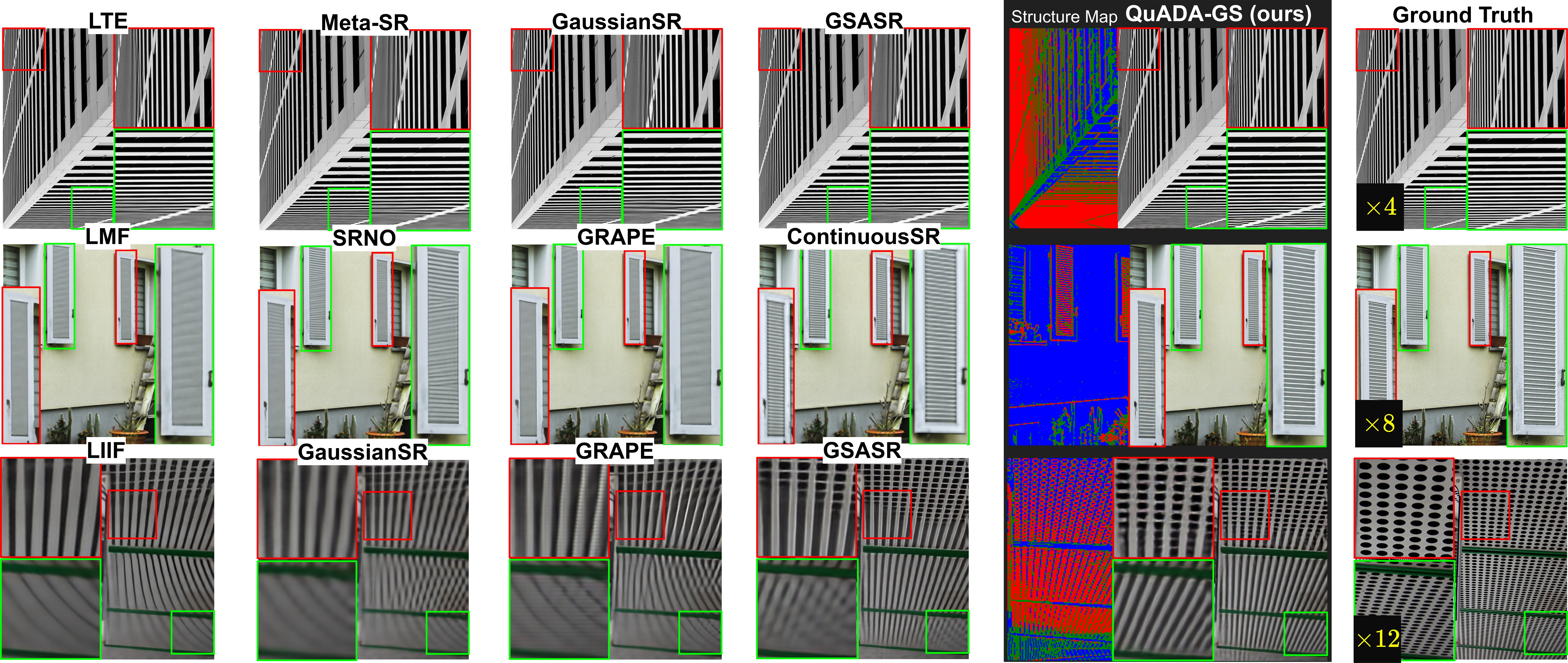}
\caption{Qualitative comparisons at $\times 4, \times 8$ and $\times 12$ between QuADA-GS and competitors, using RDN \cite{zhang2018residual} as image encoder.}
\label{fig:all_comparisons}
\end{figure*}

\textbf{Arbitrary-Scale Generalization.}
True arbitrary-scale super-resolution capability is better validated across standard benchmarks. For space, we report Urban100 (high detail density) and Set14 (low detail density) here, deferring the full suite to the Supplement. \textit{QuADA-GS excels on complex patterns, achieving reconstruction quality} (Tab. \ref{tab:urban_and_set14}) comparable to GSASR~\cite{Chen_2025_ICCV} but with a \textit{significantly lower memory footprint and inference} time (Tab. \ref{tab:comp_cost_edsr}). Our HPC modules refine exclusively high-frequency details; unlike dense methods, they are not forced to jointly process heterogeneous features. This single objective allows the network to specialize entirely on extracting rare, complex patterns from a poor low-resolution image (Fig.~\ref{fig:all_comparisons}). 

\textbf{The PSNR Penalty of Sparse Allocation.} However, \textit{lower density in flat areas inevitably penalizes PSNR: since PSNR} is based on pixel-wise MSE and weights all pixels equally, minor, perceptually negligible errors from fewer, larger Gaussians in these regions can disproportionately lower the score.

\begin{figure}[H]
\centering
\includegraphics[width=0.885\columnwidth]{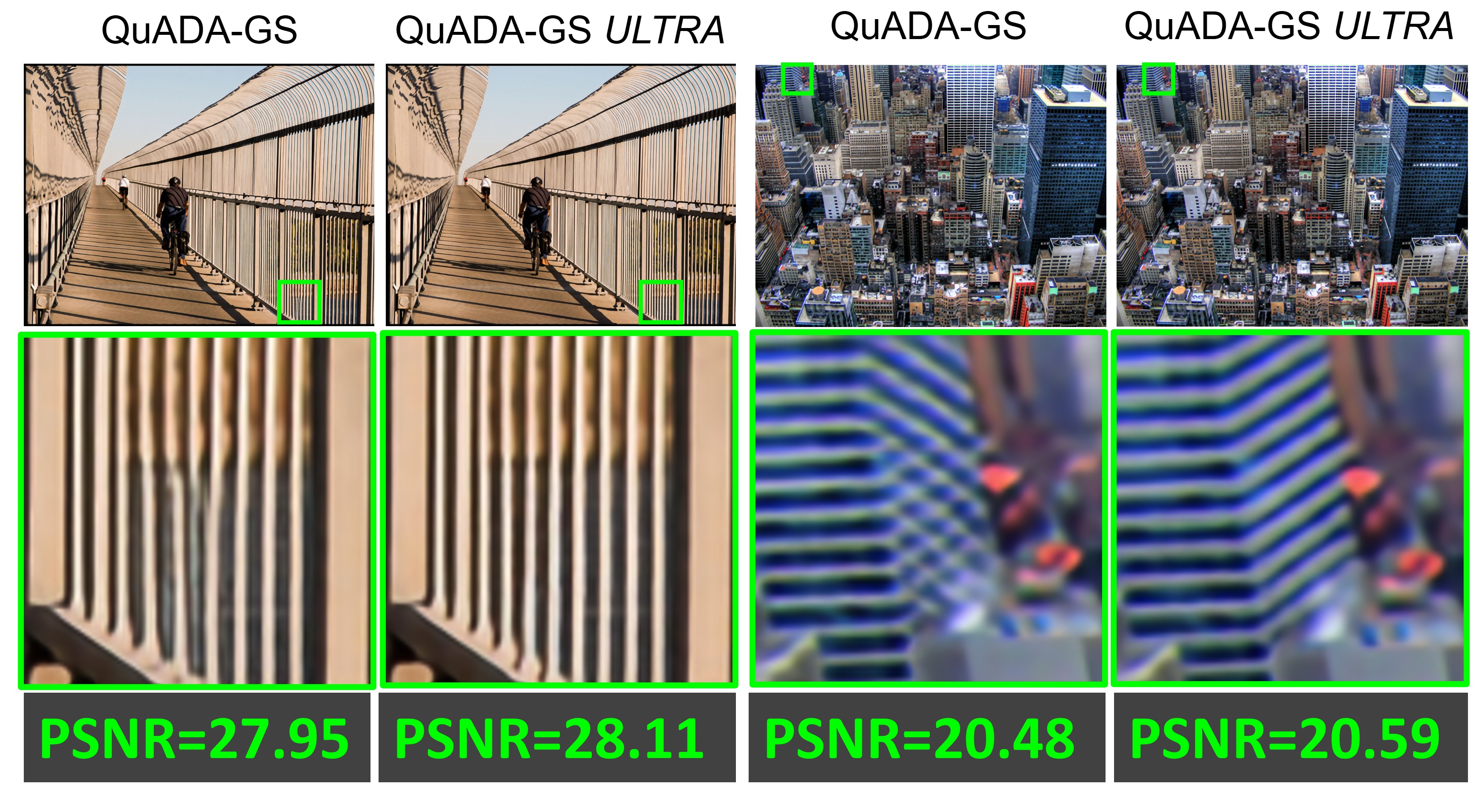}
\caption{\textbf{QuADA-GS vs. ULTRA.} ULTRA, an enhanced QuADA-GS variant, is able to captures more intricate patterns.}

\label{fig:base_vs_ultra}
\end{figure}

\textbf{Unlocking Extreme Refinement.}
To prove this lower density in flat areas causes the metric drop, we trained the \textit{ULTRA} variant. ULTRA quadruples flat-region density by uniformly upsampling the base feature map $\mathbf{F}_{base}$ to $2H\times 2W$ prior to our sparse pipeline, hitting an $8H\times 8W$ \textit{virtual} resolution. ULTRA achieves 
superior metrics across the board (Tab.~\ref{tab:urban_and_set14}) and extracts more informative patterns (Fig.~\ref{fig:base_vs_ultra}) than the base version, all while remaining faster than GSASR. Crucially, dense methods like GSASR would incur prohibitive costs if scaled to this resolution. As demonstrated in the Suppl. Mat., ULTRA uses $\sim 61\%$ less inference time and $\sim 35\%$ less memory than a dense baseline attempting to reach the same $8H\times8W$ resolution, making such extreme resolutions practical.

\newcommand{\stackpm}[2]{\begin{tabular}{@{}c@{}}#1 \\ ($\pm$ #2)\end{tabular}}
\begin{table}[t]
\centering
\resizebox{\linewidth}{!}{
\setlength{\tabcolsep}{3.5pt}
\begin{tabular}{l rrrrrr rrrrrr}
\toprule
\multirow{2}{*}{Methods} 
& \multicolumn{6}{c}{Inference Time (ms) $\downarrow$} 
& \multicolumn{6}{c}{GPU Memory (MB) $\downarrow$} \\
\cmidrule(lr){2-7} \cmidrule(lr){8-13}
& \multicolumn{1}{c}{$\times 2$} 
& \multicolumn{1}{c}{$\times 3$} 
& \multicolumn{1}{c}{$\times 4$} 
& \multicolumn{1}{c}{$\times 6$} 
& \multicolumn{1}{c}{$\times 8$} 
& \multicolumn{1}{c}{$\times 12$}
& \multicolumn{1}{c}{$\times 2$} 
& \multicolumn{1}{c}{$\times 3$} 
& \multicolumn{1}{c}{$\times 4$} 
& \multicolumn{1}{c}{$\times 6$} 
& \multicolumn{1}{c}{$\times 8$} 
& \multicolumn{1}{c}{$\times 12$} \\
\midrule
\multicolumn{13}{c}{\textit{Implicit Methods}} \\
\midrule
Meta-SR
& \RankCellTwelve{186}{5} & \RankCellTwelve{161}{6} & \RankCellTwelve{47}{2} & \RankCellTwelve{46}{2} & \RankCellTwelve{42}{2} & \RankCellTwelve{41}{2}
& \RankCellTwelve{670}{3} & \RankCellTwelve{493}{2} & \RankCellTwelve{432}{3} & \RankCellTwelve{387}{4} & \RankCellTwelve{371}{4} & \RankCellTwelve{360}{5} \\
LIIF
& \RankCellTwelve{454}{7} & \RankCellTwelve{438}{9} & \RankCellTwelve{182}{8} & \RankCellTwelve{176}{9} & \RankCellTwelve{170}{9} & \RankCellTwelve{172}{10}
& \RankCellTwelve{548}{2} & \RankCellTwelve{570}{3} & \RankCellTwelve{308}{2} & \RankCellTwelve{264}{2} & \RankCellTwelve{248}{3} & \RankCellTwelve{237}{3} \\
LTE
& \RankCellTwelve{126}{4} & \RankCellTwelve{118}{4} & \RankCellTwelve{114}{6} & \RankCellTwelve{118}{7} & \RankCellTwelve{112}{8} & \RankCellTwelve{113}{9}
& \RankCellTwelve[\textbf{490}]{490}{1} & \RankCellTwelve[\textbf{333}]{333}{1} & \RankCellTwelve[\textbf{279}]{279}{1} & \RankCellTwelve[\textbf{239}]{239}{1} & \RankCellTwelve{224}{2} & \RankCellTwelve{214}{2} \\
SRNO
& \RankCellTwelve{107}{3} & \RankCellTwelve{114}{3} & \RankCellTwelve{95}{5} & \RankCellTwelve{94}{6} & \RankCellTwelve{96}{7} & \RankCellTwelve{91}{7}
& \RankCellTwelve{6301}{10} & \RankCellTwelve{6282}{11} & \RankCellTwelve{6275}{12} & \RankCellTwelve{6270}{12} & \RankCellTwelve{6269}{12} & \RankCellTwelve{6268}{12} \\
LINF
& \RankCellTwelve{86}{2} & \RankCellTwelve{89}{2} & \RankCellTwelve{66}{3} & \RankCellTwelve{65}{4} & \RankCellTwelve{64}{4} & \RankCellTwelve{64}{5}
& \RankCellTwelve{3573}{5} & \RankCellTwelve{3412}{8} & \RankCellTwelve{3357}{9} & \RankCellTwelve{3316}{10} & \RankCellTwelve{3302}{10} & \RankCellTwelve{3292}{10} \\
LMF
& \RankCellTwelve{209}{6} & \RankCellTwelve{147}{5} & \RankCellTwelve{68}{4} & \RankCellTwelve{55}{3} & \RankCellTwelve{64}{4} & \RankCellTwelve{52}{4}
& \RankCellTwelve{4005}{6} & \RankCellTwelve{1798}{5} & \RankCellTwelve{1139}{5} & \RankCellTwelve{912}{6} & \RankCellTwelve{832}{7} & \RankCellTwelve{775}{8} \\
CiaoSR
& \RankCellTwelve{23603}{12} & \RankCellTwelve{1998}{12} & \RankCellTwelve{1165}{12} & \RankCellTwelve{716}{12} & \RankCellTwelve{616}{11} & \RankCellTwelve{540}{11}
& \RankCellTwelve{49152}{12} & \RankCellTwelve{10002}{12} & \RankCellTwelve{3331}{8} & \RankCellTwelve{1548}{8} & \RankCellTwelve{1503}{9} & \RankCellTwelve{1470}{9} \\
\midrule
\multicolumn{13}{c}{\textit{Explicit Methods}} \\
\midrule
GaussianSR
& \RankCellTwelve{754}{9} & \RankCellTwelve{717}{10} & \RankCellTwelve{686}{11} & \RankCellTwelve{692}{11} & \RankCellTwelve{666}{12} & \RankCellTwelve{688}{12}
& \RankCellTwelve{5200}{7} & \RankCellTwelve{5138}{9} & \RankCellTwelve{5048}{11} & \RankCellTwelve{5224}{11} & \RankCellTwelve{5012}{11} & \RankCellTwelve{5214}{11} \\
CSR*
& \RankCellTwelve{938}{10} & \RankCellTwelve{421}{8} & \RankCellTwelve{273}{9} & \RankCellTwelve{133}{8} & \RankCellTwelve{88}{6} & \RankCellTwelve{69}{6}
& \RankCellTwelve{6042}{9} & \RankCellTwelve{2868}{7} & \RankCellTwelve{1918}{7} & \RankCellTwelve{1070}{7} & \RankCellTwelve{771}{6} & \RankCellTwelve{559}{7} \\
GRAPE
& \RankCellTwelve[\textbf{16}]{16}{1} & \RankCellTwelve[\textbf{10}]{10}{1} & \RankCellTwelve[\textbf{8}]{8}{1} & \RankCellTwelve[\textbf{6}]{6}{1} & \RankCellTwelve[\textbf{6}]{6}{1} & \RankCellTwelve[\textbf{5}]{5}{1}
& \RankCellTwelve{2615}{4} & \RankCellTwelve{1173}{4} & \RankCellTwelve{668}{4} & \RankCellTwelve{306}{3} & \RankCellTwelve[\textbf{181}]{181}{1} & \RankCellTwelve[\textbf{90}]{90}{1} \\
GSASR
& \RankCellTwelve{1573}{11} & \RankCellTwelve{806}{11} & \RankCellTwelve{543}{10} & \RankCellTwelve{265}{10} & \RankCellTwelve{195}{10} & \RankCellTwelve{91}{7}
& \RankCellTwelve{13367}{11} & \RankCellTwelve{6000}{10} & \RankCellTwelve{3420}{10} & \RankCellTwelve{1578}{9} & \RankCellTwelve{1051}{8} & \RankCellTwelve{472}{6} \\
\begin{tabular}{@{}l@{}}\textbf{QuADA-GS} \\ \textbf{(ours)}\end{tabular}
& \RankCellTwelve[\stackpm{524}{62}]{524}{8} & \RankCellTwelve[\stackpm{251}{27}]{251}{7} & \RankCellTwelve[\stackpm{146}{14}]{146}{7} & \RankCellTwelve[\stackpm{76}{7}]{76}{5} & \RankCellTwelve[\stackpm{58}{5}]{58}{3} & \RankCellTwelve[\stackpm{44}{3}]{44}{3}
& \RankCellTwelve[\stackpm{5591}{2070}]{5591}{8} & \RankCellTwelve[\stackpm{2519}{873}]{2519}{6} & \RankCellTwelve[\stackpm{1437}{451}]{1437}{6} & \RankCellTwelve[\stackpm{698}{182}]{698}{5} & \RankCellTwelve[\stackpm{499}{113}]{499}{5} & \RankCellTwelve[\stackpm{266}{45}]{266}{4} \\
\midrule
\begin{tabular}{@{}l@{}}\textbf{QuADA-GS} \\ \textbf{ULTRA (ours)}\end{tabular}
& \stackpm{1211}{259} & \stackpm{570}{128} & \stackpm{320}{68} & \stackpm{145}{26} & \stackpm{81}{26} & \stackpm{59}{14} 
& \stackpm{32134}{8444} & \stackpm{14630}{4011} & \stackpm{8345}{2189} & \stackpm{3792}{1027} & \stackpm{2544}{692} & \stackpm{1036}{264} \\
\bottomrule
\end{tabular}
}

\caption{\textbf{Comparison of computational costs.} All models use the same EDSR backbone \cite{Lim_2017_CVPR_Workshops}. 
best results in \textbf{bold}. For our adaptive method, mean $\pm$ standard deviation is reported, as costs vary with input LR complexity.}

\label{tab:comp_cost_edsr}
\end{table}

\subsection{Ablation Study}
\label{sec:ablation}

We conduct our ablation at the representative $\times4$ scale (Tab.~\ref{tab:combined_ablation}), detailing others in the Suppl.\ Mat. Quantizing the Structure Map with thresholds $[0.1, 1.4]$ yields a superior trade-off versus full ($[0, 0]$) or no refinement ($[2, 2]$). Structure Tensor marginally outperforms Gradient Magnitude and Local Variance. Applying 1 HPC block after the first expansion and 2 after the second ($[1, 2]$) is optimal. Since Gaussian blending inherently updates features during rasterization, $[1, 2]$ provides sufficient neighbor approximation for signal integration. Relying solely on blending ($[0, 0]$) fails, while $[3, 3]$ adds costs without gains.
Replacing the EDSR baseline~\cite{Lim_2017_CVPR_Workshops} with RDN~\cite{zhang2018residual} further improves quality for a modest overhead. Tab.~\ref{tab:combined_ablation} also compares HPC against alternative sparse communications. ``Setup / Index'' denotes index creation (HPC, O-CNN, Z order curve), feature scattering (DenseScatterGatherConv), or distance computation (KNN). \textit{DenseScatterGatherConv}, inspired by scatter-based dense approaches \cite{voxelnet, pointpillars}, scatters features to a dense grid for standard convolution. However, the resulting grid contains many non-informative cells, hindering information propagation while incurring prohibitive memory costs. Conversely, following graph-based point communication \cite{pointnet, kpconv}, \textit{KNN} scales quadratically by computing all-pairs distances, quickly causing Out-Of-Memory errors. 3D sparse solutions~\cite{SubmanifoldSparseConvNet,Choy_2019_CVPR,o_cnn, octformer} target a different concept of sparsity: skipping empty space via hash tables or Morton-ordered indices, not bridging resolution levels except through approximate pooling. For example, we adapt \textit{O-CNN}~\cite{o_cnn} to our setting, but its tree sorting incurs high memory overhead, while its lack of cross-level communication and uniform maximum depth isolate coarse features in empty blocks, causing kernels to sample empty space and degrade quality; fixing $Z{=}0$ further wastes volumetric queries. A \textit{Z-order curve} baseline builds a slightly faster and lighter index (even if our method enables faster search), but linearizing the 2D space into a 1D representation may map topologically close points to distant locations, again lowering quality. Finally, HPC search time remains constant ($\sim 0.47$\,ms), confirming its $\mathcal{O}(1)$ complexity.

\begin{table}[b]
\centering

\resizebox{\columnwidth}{!}{
\begin{tabular}{l cccc}
\toprule
\textbf{Configuration} & \textbf{Time $\downarrow$} & \textbf{Mem $\downarrow$} & \textbf{PSNR $\uparrow$} & \textbf{SSIM $\uparrow$} \\
\midrule

\rowcolor{gray!15}
\multicolumn{5}{l}{\textit{Quantization Threshold for the Structure Map $[\phi_1, \phi_2]$.}} \\
$[0, 0]$ Each feature is refined up to level 2 & 
\RankCellThree[$275$]{275}{3} & \RankCellThree[$5921$]{5921}{3} & \RankCellThree[$31.74$]{31.74}{1} & \RankCellThree[$0.8521$]{0.8521}{1} \\
\textbf{$\mathbf{[0.1, 1.4]}$ Mixed levels} & 
\RankCellThree[$146$]{146}{2} & \RankCellThree[$1437$]{1437}{2} & \RankCellThree[$31.70$]{31.70}{2} & \RankCellThree[$0.8514$]{0.8514}{2} \\
$[2, 2]$ No features are refined. & 
\RankCellThree[\textbf{$129$}]{129}{1} & \RankCellThree[\textbf{$663$}]{663}{1} & \RankCellThree[$31.00$]{31.00}{3} & \RankCellThree[$0.8384$]{0.8384}{3} \\

\midrule
\rowcolor{gray!15}
\multicolumn{5}{l}{\textit{Important weights $[w_{texture}, w_{edge}]$ for the per-pixel score.}} \\
$[1, 1]$ Equal Importance & 
\RankCellFive[$147.84$]{147}{3} & \RankCellFive[$1506$]{1506}{4} & \RankCellFive[$31.6499$]{31.6499}{5} & \RankCellFive[$0.8514$]{0.8514}{1} \\
$[1, 2]$ Edge-Favored & 
\RankCellFive[$148.90$]{148}{5} & \RankCellFive[$1522$]{1522}{5} & \RankCellFive[$31.6535$]{31.6535}{4} & \RankCellFive[$0.8514$]{0.8514}{1} \\
$[2, 1]$ Texture-Favored & 
\RankCellFive[$148.03$]{0.6}{4} & \RankCellFive[$1472$]{1472}{3} & \RankCellFive[$31.6742$]{31.6742}{3} & \RankCellFive[$0.8514$]{0.8514}{1} \\
$\mathbf{[3, 1]}$\textbf{ Strongly Texture-Favored} & 
\RankCellFive[\textbf{$146.08$}]{146.08}{1} & \RankCellFive[\textbf{$1437$}]{1437}{1} & \RankCellFive[\textbf{$31.7010$}]{31.7010}{2} & \RankCellFive[$0.8514$]{0.8514}{1} \\
$[4, 1]$ Extremely Texture-Favored & 
\RankCellFive[$146.65$]{146.65}{2} & \RankCellFive[$1443$]{1443}{2} & \RankCellFive[\textbf{$31.6998$}]{31.6998}{1} & \RankCellFive[\textbf{$0.8514$}]{0.8514}{1} \\

\midrule
\rowcolor{gray!15}
\multicolumn{5}{l}{\textit{Complexity Measure for the Score Map.}} \\
\textbf{Structure Tensor}& 
$-$ & $-$ & \RankCellThree[\textbf{$31.7010$}]{31.7010}{1} & \RankCellThree[\textbf{$0.8514$}]{0.8514}{1} \\
Gradient Magnitude & 
$-$ & $-$ & \RankCellThree[$31.6965$]{31.6965}{3} & \RankCellThree[$0.8507$]{0.8507}{3} \\
Local Variance & 
$-$ & $-$ & \RankCellThree[$31.6986$]{31.6986}{2} & \RankCellThree[$0.8509$]{0.8509}{2} \\

\midrule
\rowcolor{gray!15}
\multicolumn{5}{l}{\textit{Number of HPC Blocks after Each Blind Expansion $[N_1, N_2]$.}} \\
$[0, 0]$ No interaction between features & 
\RankCellSix[\textbf{$136$}]{136.97}{1} & \RankCellSix[\textbf{$846$}]{846.58}{1} & \RankCellSix[$30.0411$]{30.0411}{6} & \RankCellSix[$0.8198$]{0.8198}{6} \\
$[2, 0]$ Two only after first Blind Exp. & 
\RankCellSix[$140$]{140.58}{2} & \RankCellSix[$1229$]{1229}{2} & \RankCellSix[$30.9137$]{30.9137}{5} & \RankCellSix[$0.8366$]{0.8366}{5} \\
$[0, 2]$ Two only after second Blind Exp. & 
\RankCellSix[$143$]{143.78}{3} & \RankCellSix[$1433$]{1433}{4} & \RankCellSix[$31.3619$]{31.3619}{4} & \RankCellSix[$0.8443$]{0.8443}{4} \\
$[1, 1]$ One for each Blind Exp. & 
\RankCellSix[$144$]{144}{4} & \RankCellSix[$1428$]{1428}{3} & \RankCellSix[$31.5374$]{31.5374}{3} & \RankCellSix[$0.8492$]{0.8492}{3} \\
$\mathbf{[1, 2]}$ \textbf{One for the first, two for the second }& 
\RankCellSix[$146$]{146}{5} & \RankCellSix[$1437$]{1437}{5} & \RankCellSix[$31.7010$]{31.7010}{2} & \RankCellSix[$0.8514$]{0.8514}{2} \\
$[3, 3]$ Three for each Blind Exp. & 
\RankCellSix[\textbf{$155$}]{155}{6} & \RankCellSix[\textbf{$1446$}]{1446}{6} & \RankCellSix[\textbf{$31.7082$}]{31.7082}{1} & \RankCellSix[\textbf{$0.8521$}]{0.8521}{1} \\

\midrule
\rowcolor{gray!15}
\multicolumn{5}{l}{\textit{The choice of feature extractor.}}\\
EDSR\_baseline & 
\RankCellTwo[\textbf{$\mathbf{146}$}]{146}{1} & \RankCellTwo[{$\mathbf{1437}$}]{1437}{1} & \RankCellTwo[$31.70$]{31.70}{2} & \RankCellTwo[$0.8515$]{0.8514}{2} \\
RDN & 
\RankCellTwo[$166$]{166}{2} & \RankCellTwo[$1544$]{1544}{2} & \RankCellTwo[$\mathbf{31.92}$]{31.92}{1} & \RankCellTwo[\textbf{$\mathbf{0.8551}$}]{0.8551}{1} \\

\bottomrule
\end{tabular}
}

\vspace{0.3cm} 

\resizebox{\columnwidth}{!}{%
\begin{tabular}{l c c c c c c}
\toprule

\multirow{2}{*}{\textbf{Method}}
& \textbf{Setup / Index} & \textbf{Setup / Index} & \textbf{Search} & \textbf{Communication} & \multirow{2}{*}{\textbf{PSNR $\uparrow$}} & \multirow{2}{*}{\textbf{SSIM $\uparrow$}} \\
& \textbf{T $\downarrow$} & \textbf{M $\downarrow$} & \textbf{T $\downarrow$} & \textbf{M $\downarrow$} & & \\

\midrule
\rowcolor{gray!15}
\multicolumn{7}{l}{\textit{Alternatives to our HPC.}} \\

\textbf{HPC (ours)}
& \RankCellFive[$5.14/4.99/2.38$]{5.14/4.99/2.38}{2}
& \RankCellFive[$104/28/8$]{104/28/8}{2}
& \RankCellFive[\textbf{$\mathbf{0.46/0.48/0.47}$}]{0.46/0.48/0.47}{1}
& \RankCellFive[\textbf{$\mathbf{3289/788/228}$}]{3289/788/228}{1}
& \RankCellFive[$\mathbf{37.26/31.70/28.08}$]{37.26/31.70/28.08}{1}
& \RankCellFive[$\mathbf{0.9511/0.8515/0.7375}$]{0.9511/0.8515/0.7375}{1}
\\

DenseScatterGatherConv
& \RankCellFive[$20.3/9.73/5.66$]{20.3/9.73/5.66}{4}
& \RankCellFive[$2613/668/196$]{2613/668/196}{4}
& --
& \RankCellFive[$4272/1070/306$]{4272/1070/306}{4}
& \RankCellFive[36.92/31.38/27.71]{37.02/31.48/27.71}{3}
& \RankCellFive[0.9508/0.8489/0.7333]{0.9521/0.8521/0.7382}{2}
\\

KNN
& \RankCellFive[OOM/OOM/30.98]{OOM/OOM/30.98}{5}
& \RankCellFive[\quad--\quad/\quad--\quad/3980]{--/--/3980}{5}
& --
& \RankCellFive[\quad--\quad/\quad--\quad/228]{--/--/228}{5}
& \RankCellFive[\quad--\quad/\quad--\quad/28.06]{--/--/28.06}{5}
& \RankCellFive[\quad--\quad/\quad--\quad/0.7379]{--/--/0.7379}{5}
\\

O-CNN\textbf{*}
& \RankCellFive[\textbf{$17.18/12.07/10.04$}]{17.18/12.07/10.04}{3}
& \RankCellFive[\textbf{$2418/614/184$}]{2418/614/184}{3}
& \RankCellFive[$2.69/1.19/0.65$]{2.69/1.19/0.65}{5}
& \RankCellFive[$3814/1710/1277$]{3814/1710/1277}{2}
& \RankCellFive[36.88/31.26/27.54]{36.88/31.26/27.54}{4}
& \RankCellFive[0.9432/0.8468/0.7322]{0.9432/0.8468/0.7322}{4}
\\

Z-order Curve
& \RankCellFive[\textbf{$\mathbf{2.04/2.62/1.34}$}]{2.04/2.62/1.34}{1}
& \RankCellFive[\textbf{$\mathbf{37.7/12.7/2.03}$}]{37.7/12.7/2.03}{1}
& \RankCellFive[$0.80/1.1/0.94$]{2.69/1.19/0.65}{3}
& \RankCellFive[\textbf{$\mathbf{3289/788/228}$}]{3289/788/228}{1}
& \RankCellFive[36.98/31.28/27.78]{36.98/31.28/27.78}{2}
& \RankCellFive[0.9478/0.8476/0.7324]{0.9478/0.8476/0.7324}{3}
\\

\bottomrule
\end{tabular}%
}


\caption{\textbf{Ablation studies.} \textit{Top:} Methodological components at $\times4$; \textbf{bold} marks our final configuration. \textit{Bottom:} HPC vs. sparse communication alternatives in time $T$ (ms) and memory $M$ (MB) across $\times2/\times4/\times8$; best results in \textbf{bold}. \textbf{*} Adapted to our setting.}

\label{tab:combined_ablation}
\end{table}

\section{Conclusion}

We introduced QuADA-GS, a highly efficient GS-based method for ASR that leverages dynamic densification to bypass the prohibitive costs of dense architectures. By adaptively allocating Gaussians strictly where structural complexity demands it, our approach effectively extracts high-frequency patterns and achieves an optimal quality-cost trade-off. To push fidelity further, our \textit{ULTRA} variant achieves superior metrics at a fraction of the cost of dense equivalents. A current \textbf{limitation} is the empirical tuning of the Structure Map's quantization thresholds on DIV2K. While benchmarks demonstrate robust generalization, future work will explore fully learned quantization.
\clearpage

{
    \small
    \bibliographystyle{ieeenat_fullname}
    \bibliography{main}
}

\end{document}